\documentclass{article} %
\usepackage[final]{neurips_2025}

\usepackage[utf8]{inputenc} %
\usepackage[T1]{fontenc}    %
\usepackage{hyperref}       %
\usepackage{url}            %
\usepackage{booktabs}       %
\usepackage{nicefrac}       %
\usepackage{microtype}      %
\usepackage{xcolor}         %
\usepackage{wrapfig}
\usepackage{lipsum}
\usepackage{makecell}
\usepackage{multicol}
\usepackage{multirow}

\usepackage{subcaption}

\usepackage{graphicx} %
\usepackage{amsmath}
\usepackage{amsfonts}
\usepackage{amssymb}
\usepackage{amsthm}

\theoremstyle{plain}
\newtheorem{theorem}{Theorem}[section]
\newtheorem*{theorem*}{Theorem}

\newtheorem*{lemma*}{Lemma}
\newtheorem{corollary}[theorem]{Corollary}
\theoremstyle{definition}

\theoremstyle{remark}

\usepackage{algorithm}
\usepackage{algorithmic}

\title{Prompt Tuning Decision Transformers with Structured and Scalable Bandits}

\author{Finn Rietz\thanks{Work performed while the author was an intern at King (part of Microsoft Gaming).} \\
\"Orebro University\\
\texttt{finn.rietz@oru.se} \\
\And
Oleg Smirnov\\
King AI Labs, Microsoft Gaming \\
\texttt{oleg.smirnov@microsoft.com}
\AND
Sara Karimi\\
King AI Labs, Microsoft Gaming \\
\texttt{sara.karimi@king.com}
\And
Lele Cao \\
King AI Labs, Microsoft Gaming \\
\texttt{lelecao@microsoft.com}
}

\begin{document}

\maketitle

\begin{abstract}
Prompt tuning has emerged as a key technique for adapting large pre-trained Decision Transformers (DTs) in offline Reinforcement Learning (RL), particularly in multi-task and few-shot settings. The Prompting Decision Transformer (PDT) enables task generalization via trajectory prompts sampled uniformly from expert demonstrations -- without accounting for prompt informativeness.
In this work, we propose a bandit-based prompt-tuning method that learns to construct optimal trajectory prompts from demonstration data at inference time. 
We devise a structured bandit architecture operating in the trajectory prompt space, achieving linear rather than combinatorial scaling with prompt size.  Additionally, we show that the pre-trained PDT itself can serve as a powerful feature extractor for the bandit, enabling efficient reward modeling across various environments.
We theoretically establish regret bounds and demonstrate empirically that our method consistently enhances performance across a wide range of tasks, high-dimensional environments, and out-of-distribution scenarios, outperforming existing baselines in prompt tuning.
\end{abstract}

\section{Introduction}
\label{sec:introduction}

The ability to exploit large amounts of offline data is crucial for training foundation models capable of generalizing across diverse tasks~\citep{radford2018improving, reed2022generalist, brohan2022rt}. In Reinforcement Learning (RL), a seminal contribution is the Decision Transformer (DT)~\citep{chen2021decision}, which reframes offline RL~\citep{levine2020offline} as a sequence modeling problem, thereby unlocking powerful Transformer architectures for offline RL. DT is particularly well-suited for offline RL because it sidesteps the well-known instabilities of temporal-difference learning with function approximation~\citep{sutton_rlbook}, which are exacerbated under distribution shift in the offline setting, by replacing discounted, moving-target value estimates with stationary return-to-go conditioning, autoregressive sequence modeling, and causal attention mechanisms~\citep{chen2021decision}. The Prompting Decision Transformer (PDT)~\citep{xu2022prompting} further extends DT from single-task to multi-task settings, enabling large-scale models and generalized pre-training in offline multi-task and few-shot RL~\citep{xu2022prompting, mitchell2021offline}. Analogous to prompting in Large Language Models (LLMs), PDT differentiates tasks through a \textit{stochastic trajectory prompt} prepended to the context, allowing it to identify and model optimal action marginals for each task in the offline dataset.
This makes PDT particularly appealing for offline multi-task RL, as it avoids additional learning on the downstream task 
while enabling efficient and robust adaption of the model's behavior at inference time, solely through adjustments of the prompt.

However, PDT samples prompts uniformly, overlooking that not all prompts are equally informative, even in fully observable MDP.
We hypothesize that the sampling of non-informative prompts from expert demonstrations can diminish PDT's ability to differentiate between tasks, thereby leading to performance degradation.

While improving on PDT's uninformed prompt-sampling strategy, prior works on PDT prompt-tuning suffer from key limitations.
The approach by~\citet{yuan2024pre} replaces trajectory prompts with less-expressive goal-conditioning and relies on hindsight relabeling. The generative approaches by~\citet{hu2023prompt, hu2024prompt} are not applicable in discrete settings and don't adhere to causal relationships between prompt tokens. Furthermore, all of these works treat prompts as flat, unstructured inputs and operate directly on MDP modalities, which leads to poor scaling with prompt size and state- and action-space sizes. 

To address these shortcomings, we introduce a scalable, robust, and computationally efficient bandit-based prompt-tuning method for PDT. 
By exploiting the structure of the prompt space, our method reduces the complexity of prompt selection from combinatorial to linear in the prompt size. 
Moreover, we show how to leverage the pre-trained PDT as a feature extractor to obtain compact prompt representations that enable efficient deployment even in high-dimensional (e.g. pixle-based) settings.
By optimizing prompt selection, our approach boosts downstream task performance without costly weight updates to the underlying Transformer backbone. 
Our experiments reveal clear performance gains with the proposed method, effectively bridging the gap between pre-training and adaptation.

\section{Preliminaries}
\label{sec:preliminaries}
In this section, we introduce key concepts and terminologies that form the foundation of this work.

\subsection{Problem definition: Offline multi-task RL}
The offline multi-task RL problem is formalized similarly to prior works~\citep{xu2022prompting, mitchell2021offline}. The objective is to solve a set of training tasks $\mathcal{T}^\text{train}$, with the option to evaluate task generalization capabilities on a holdout test set $\mathcal{T}^\text{test}$. 

For each task $\mathcal{T}_i \in \mathcal{T}^\text{train}$, a dataset $\mathcal{D}_i$ is provided, consisting of trajectories sampled from the corresponding MDP $\mathcal{M}_i = \langle \mathcal{S}_i, \mathcal{A}_i, r_i, d_i, \gamma_i, \mu_i^0 \rangle$. Here, $\mathcal{S}_i$ is the state space, $\mathcal{A}_i$ is the action space, $r_i: \mathcal{S}_i \times \mathcal{A}_i \to \mathbb{R}$ represents the reward function, $d_i: \mathcal{S}_i \times \mathcal{A}_i \times \mathcal{S}_i \to [0, 1]$ defines the discrete-time transition dynamics, $\gamma_i \in (0, 1]$ is the discount factor, and $\mu_i^0$ is the initial state distribution of MDP $i$. 

The goal is to learn a generalized policy, $\pi(\mathbf{s}, \psi) \to \mathbf{a}$, capable of solving all tasks in $\mathcal{T}^\text{train}$. 
Here, $\psi$ serves as an auxiliary input to the policy that sufficiently describes a task from $\mathcal{T}$. 
In the case of PDT, $\psi$ is the \textit{stochastic trajectory prompt} sampled from the demonstration set for the target task.
For each task, the optimal generalized policy is expected to maximize the corresponding expected discounted reward objective specific to task $i$.
\begin{equation}\label{eq:generalized-objective}
    J(\pi, i) = \mathbb{E} \left[ \sum_{t=0}^\infty \gamma_{i}^t r_i(\mathbf{s}_t, \mathbf{a}_t) \right]
\end{equation}
The expectation is taken over trajectories where actions are sampled from the policy $\pi$, and states are sampled using the transition dynamics $d_i$ of the MDP $\mathcal{M}_i$. 
In the offline setting, only the pre-collected datasets $\mathcal{D} = \{\mathcal{D}_0, \dots, \mathcal{D}_n\}$ are available for learning, with no additional data collection allowed. 
A simulator may be used for policy evaluation but not for gathering new training data due to the offline nature of the problem.

\subsection{Prompting Decision Transformer}
\citet{chen2021decision} reframed RL as a sequence modeling problem, leveraging the capabilities of Transformer architectures to model trajectories. A trajectory in DT is represented as a sequence of triplets $(\hat{r}_t, \mathbf{s}_t, \mathbf{a}_t)$, where $\hat{r}_t = \sum_{t' = t} ^T r_{t'}$ is the return-to-go, $\mathbf{s}_t$ is the state, and $\mathbf{a}_t$ is the action at time step $t$. DT captures the temporal and causal dependencies between states, actions, and rewards, enabling the model to predict optimal actions directly, without relying on explicit value functions or policies. 

\citet{xu2022prompting} extended DT to a multi-task offline RL setting by introducing stochastic trajectory prompts as task-specific context, enabling the model to identify underlying MDPs.  
Such a prompt $\rho$ is composed of $J$ trajectory segments, each of length $H$, resulting in a total of $J \times H \times 3$ prompt tokens, as per Equation~\ref{eq:pdt}, where the superscript $^\star$ denotes tokens associated with the prompt.

\begin{align}
    \rho &= \big(
        \overbrace{
        \hat{r}_j^\star, \mathbf{s}_j^\star, \mathbf{a}_j^\star, 
        \dots,
        \hat{r}_{j+H}^\star, \mathbf{s}_{j+H}^\star, \mathbf{a}_{j+H}^\star}
        ^{\text{ $\Tilde{\tau}_1$: segment 1}},
        \dots,
        \overbrace{
        \hat{r}_k^\star, \mathbf{s}_k^\star, \mathbf{a}_k^\star, 
        \dots,
        \hat{r}_{k+H}^\star, \mathbf{s}_{k+H}^\star, \mathbf{a}_{k+H}^\star}
        ^{\text{$\Tilde{\tau}_J$: segment } J}
    \big)
    \label{eq:pdt} \\
    \mathbf{x} &= 
        \big( \rho \big) \odot
        \big(
        \overbrace{
        \hat{r}_{t-L}, \mathbf{s}_{t-L}, \mathbf{a}_{t-L}, 
        \hat{r}_{t-L+1}, \mathbf{s}_{t-L+1}, \mathbf{a}_{t-L+1}
        \dots,
        \hat{r}_{t}, \mathbf{s}_{t}, \mathbf{a}_{t}
        }^{\omega_{L:t}: L \text{ most recent transitions}}
        \big)
    \label{eq:pdt-seq}
\end{align}

For a particular training task $\mathcal{T}_i \in \mathcal{T}^\text{train}$, PDT learns to model the sequence in Equation~\ref{eq:pdt-seq} by autoregressively predicting the action tokens, where $\odot$ denotes concatenation. 
While PDT randomly samples the $J$ segments that constitute the prompt $\rho$ from a set of expert demonstrations $\mathcal{P}$, we propose to optimize segment composition to enhance the downstream task performance with a Multi-Armed Bandit approach.

\subsection{Multi-Armed Bandits}
Multi-Armed Bandits (MABs) provide a framework for optimizing stochastic reward functions, making them an effective tool for tasks like prompt tuning. 
In the standard MAB setting, at each time step $k$, the agent selects an action (or ``arm'') $a_k \in \mathcal{A}$, where $\mathcal{A}$ is the set of available arms. 
Here, we use $k$ to denote bandit time steps, to avoid confusion with the MDPs time $t$.
The agent then receives a stochastic reward $r_k \sim R(a_k)$, with the goal of maximizing the cumulative reward $\sum_{k=1}^K r_k$ over a time horizon $K$~\citep{auer2002finite}. This requires balancing the exploration of arms to gather information about their reward distributions $R(a)$ and exploitation of arms with known high rewards while minimizing cumulative regret, defined as:
\begin{equation}
    \mathrm{Regret}(K) = \sum_{k=1}^K \left[ \max_{a \in \mathcal{A}} \mathbb{E}[R(a)] - \mathbb{E}[r_k] \right]
\end{equation}

Contextual MABs (CMAB) extend this framework by incorporating side information (or ``context'') $\mathbf{c}_k \in \mathcal{C}$ which is observed before selecting an arm. The reward distribution is then conditioned on both the arm and the context, $r_k \sim R(a_k \mid \mathbf{c}_k)$. The agent's objective is to learn a policy $\pi: \mathcal{C} \to \mathcal{A}$ that maximizes the expected reward $\mathbb{E}\left[\sum_{k=1}^K R(\pi(\mathbf{c}_k) \mid \mathbf{c}_k)\right]$. By leveraging shared features across arms through the context $\mathbf{c}_k$, CMABs enable more efficient learning and better generalization, particularly in settings where arms share intrinsic characteristics~\citep{li2010contextual}.

\section{Related Work}
\label{sec:related_work}
Recently, there has been a surge in methods across various domains aimed at enhancing the performance and generalization of Transformer-based approaches through automatic prompt tuning. 

\textbf{Prompting in LLMs}. For LLMs, \citet{pmlr-v235-chen24e} introduced {I}nstruct{Z}ero, which uses Bayesian optimization to explore low-dimensional soft prompt vectors. These vectors are then passed to an open-source LLM to generate instructions for the black-box LLM. Building on this, \citet{lin2023use} proposed INSTINCT, which replaces the Gaussian process in Bayesian optimization with a neural network surrogate, leveraging a neural bandit algorithm to enhance expressivity. Additionally, \citet{shi2024best} demonstrated that the fixed-budget MAB framework enables learning the optimal prompt within a limited number of LLM evaluations. These methods rely on optimization in continuous spaces, followed by prompt reconstruction, whereas our approach directly operates within the PDT prompt space.

\textbf{Multi-task DT}. For RL, \citet{hu2023prompt} proposed generating prompt candidates for PDT by perturbing initial trajectories with Gaussian noise and applying online or offline feedback to guide a ranking-based optimization. Prompt Diffuser~\citep{hu2024prompt} similarly frames instruction optimization as a conditional generative modeling task, synthesizing prompts from random noise. In contrast, our method constructs prompts directly from expert demonstrations, which is particularly beneficial in settings like discrete state and action spaces, where noise-based approaches are less effective.

\citet{wang2024hierarchical} introduced hierarchical prompting, using two levels of context: one for encoding task-specific information and another for guiding rollouts via demonstration segments. Hyper-Decision Transformer~\citep{xu2023hyper} augments DT with a hyper-network for adaptation to novel tasks, but both methods rely on large quantities of demonstration data. Our method, by contrast, selects high-quality prompts from only a few expert trajectories. 
Other approaches focus on representation learning or adaptation, e.g., \citet{wang2024meta} focus on disentangled world models, while \citet{xie2023future} propose latent-variable conditioned RL and fine-tuning.
Our approach avoids updating the transformer backbone and relies solely on test-time prompt optimization.

Finally, MGPO~\citep{yuan2024pre} proposes a general framework for online prompt tuning. While sharing a similar goal, we provide a method to identify the best prompt from a demonstration dataset, while MGPO iteratively refines the prompt with data collected during online rollouts.

\textbf{LLM for RL}. At the intersection of LLM and RL domains, \citet{yang2024pre} and the closely related LaMo method~\citep{shi2024unleashing} proposed leveraging a pre-trained LLM as an initializer for PDT, harnessing rich linguistic knowledge to boost performance on unseen tasks. In another work, \citet{zhengdecomposed} introduced an approach to decompose the prompt into cross-task and task-specific components, ensuring more robust test-time adaptation to unseen tasks. Furthermore, the model is initialized by incorporating parameters from a pre-trained language model to provide it with prior knowledge. Compared to those, our approach avoids reliance on a dedicated LLM, sidestepping the fine-tuning and scalability challenges inherent in such methods.

\section{Method}
\label{sec:method}
We now present our inference time prompt-tuning bandit method for offline multi-task RL. Given a dataset $\mathcal{D} = \{ \mathcal{D}_0, \dots, \mathcal{D}_n \}$ of trajectories for $n$ training tasks $\mathcal{T}^\text{train}$, we utilize the original PDT method to learn the generalized policy described in Equation~\ref{eq:generalized-objective}. Details of the PDT algorithm and training process are provided by~\citet{xu2022prompting}.

An optimal PDT model $\pi^*(\mathbf{x}; \theta)$ is assumed to be trained until convergence on $\mathcal{D}$. After training, the goal is to evaluate and enhance the model's performance on a specific task $\mathcal{T}_i$, either from the set of holdout test tasks $\mathcal{T}^\text{test}$ or from the training set $\mathcal{T}^\text{train}$. We assume access to a small set of demonstrations $\mathcal{P}_i$ for the target task, which serve as a source for sampling prompts, as well as a simulator of the corresponding $\mathcal{M}_i$ to perform online evaluations. 
Our primary objective is to \textit{identify the stochastic trajectory prompt constructible from $\mathcal{P}_i$ that maximizes performance} when applied to the pre-trained generalized policy.

\subsection{Bandit-based Prompt Tuning}

\begin{figure*}[ht]%
    \centering
    \includegraphics[width=0.9\linewidth]{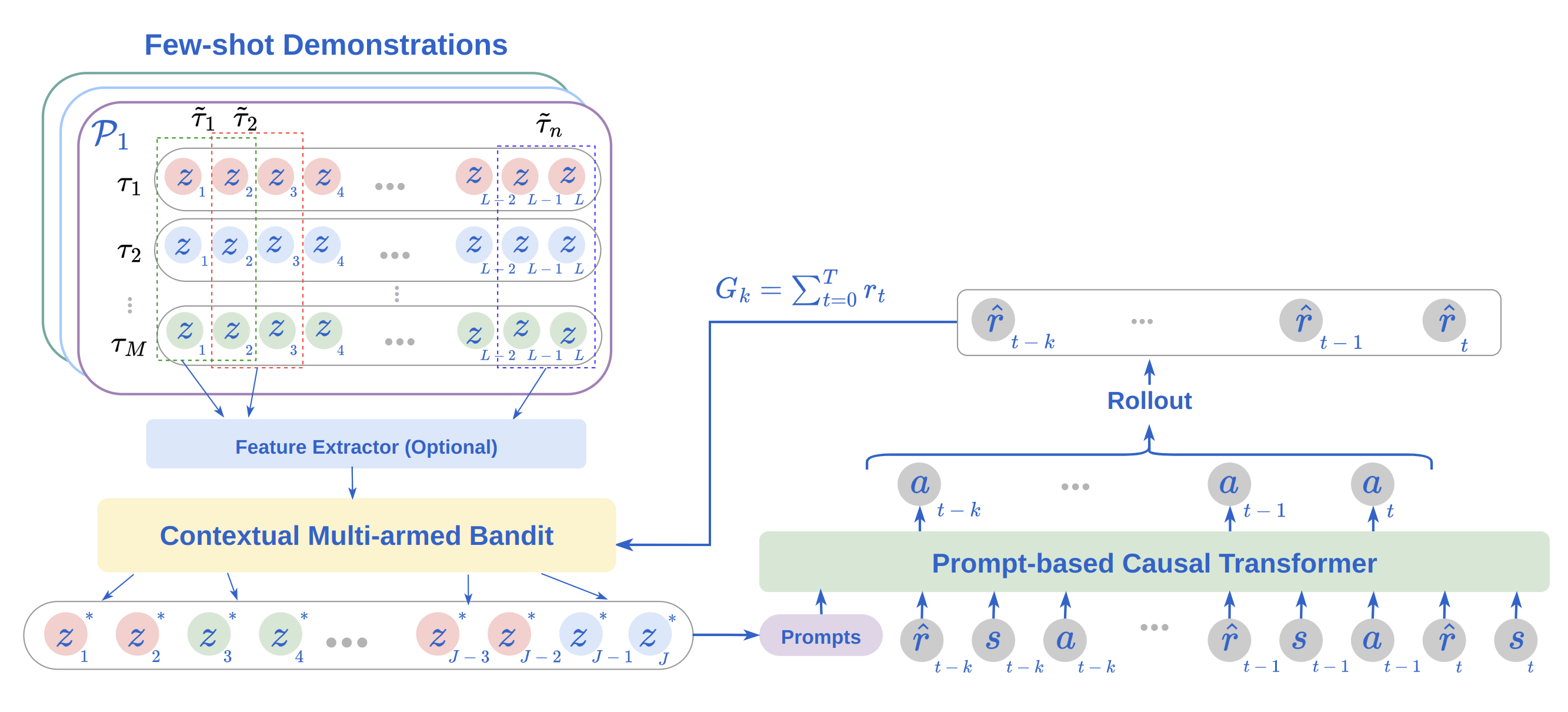}
    \caption{Overview of our bandit-based prompt-tuning method for multi-task learning with PDT. Each $z_i$ represents a triplet $(\hat{r}_i, \mathbf{s}_i, \mathbf{a}_i)$, each $\tilde{\tau}$ represents a prompt segment, each $\tau$ represents a demonstration trajectory. The bandit explores the demonstration dataset $\mathcal{P}_i$ for the current task $i$ to find the best prompt $\rho^* = (\tilde{\tau}_1^*, \dots, \tilde{\tau}_J^*)$. The online return $G_k$ achieved by the underlying PDT model at round $k$ and using prompt $\rho_k$ serves as a reward for the bandit. 
    }
    \label{fig:prompt_dt_bandit}
\end{figure*}

We propose leveraging a bandit-based approach to efficiently identify the optimal trajectory prompt. A na\"ive implementation of bandit-based prompt tuning would involve deploying a bandit with one arm for each possible prompt $\rho$ constructible from $\mathcal{P}_i$, i.e. considering all possible combinations of segments among demonstration trajectories. This approach, however, scales extremely poorly, as the number of prompts (and consequently, the size of the bandit problem) grows linearly with the total number of transitions in $\mathcal{P}_i$ and combinatorially with $J$, the number of segments in each prompt. Additionally, treating each prompt as a separate and independent arm disregards prompt similarity, leading to inefficient sample complexity. Thus, we propose to optimize the selection of segments $\Tilde{\tau}$ that form the stochastic trajectory prompt $\rho = (\Tilde{\tau}_1, \dots, \Tilde{\tau}_J)$ with a special \textit{contextual} MAB algorithm.

As illustrated in Figure~\ref{fig:prompt_dt_bandit}, at each round $k$, the bandit selects a prompt $\rho_k$ 
consisting of $J$ segments, balancing exploration and exploitation in the prompt space.
The process of constructing the prompt $\rho_k$ through the bandit mechanism is discussed in detail in the next section.
We then perform the $k$-th rollout of the PDT $\pi^*(\mathbf{x}; \theta)$  in $\mathcal{M}_i$, using the selected prompt $\rho_k$ and the most recent transitions, as per Equation~\ref{eq:pdt-seq}.
The resulting performance $G_i^k = \sum_{t=0}^T r_i(\mathbf{s}_t, \mathbf{a}_t) \mid \mathbf{a}_t \sim \pi^*(\mathbf{x}_k; \theta)$ serves as reward from the bandit's perspective and is stored together with the prompt $\rho_k$ for training the bandit. The algorithmic pseudocode for prompt-tuning with our bandit is provided in the Algorithm~\ref{alg:bandit_steps} in Appendix~\ref{app:algorithms}.

\subsection{Scalable and Sample-Efficient Bandit Architecture}
To address the aforementioned scalability issues in a na\"ive bandit, we design a structured bandit architecture that factorizes the problem across prompt segments.
Our structured architecture features $J$ arms, one for each segment in the stochastic trajectory prompt. Each arm $j \in \{1, \dots, J \}$ maintains an \textit{independent} reward model $\phi_j$, which predicts the performance of the underlying PDT when a given segment $\Tilde{\tau}$ is placed at position $j$ in the prompt. This decomposition reduces the complexity of the search space from combinatorial to linear in $J$.
Exploration can be performed using strategies such as $\epsilon$-greedy, Upper Confidence Bounds (UCB)~\citep{li2010contextual, zhou2020neural}, Thompson Sampling (TS)~\citep{thompson1933likelihood}, or other mechanisms.

To select the prompt $\rho_k$, each model $\phi_j$ predicts the reward for each segment $\{ \Tilde{\tau}_0, \dots, \Tilde{\tau}_n \} \in \mathcal{P}_i$\footnote{
Note that if $\mathcal{P}_i$ contains $M$ expert trajectories, each of length $L$, and each prompt segment has a length $H$, the total number of segments is given by $|\mathcal{P}_i| = M \times (L - H + 1)$.} in its position, $j$, in the prompt, resulting in a prediction matrix $\mathbf{Y} \in \mathcal{R}^{J \times |\mathcal{P}_i|}$.
The $J$ segments with the highest predicted performance are then selected by computing $\arg \max$ over the segment dimension of $\mathbf{Y}$, subject to the chosen exploration strategy. 
This yields the prompt $\rho_k = (\tilde{\tau}_1^k, \dots, \tilde{\tau}_J^k)$.
After the rollout $k$ of the PDT with prompt $\rho_k$ in $\mathcal{M}_i$, all reward models are independently updated on their accumulated $\langle \tilde{\tau}_j^k, G_i^k \rangle$ pairs.
The complete algorithmic pseudocode for prompt selection and update with our structured bandit architecture is provided in Algorithms~\ref{alg:select_prompt} and \ref{alg:update_bandit} in Appendix~\ref{app:algorithms}.

\subsection{Regret Analysis}
We present a regret bound for our bandit architecture, assuming that the reward for a prompt $\rho = (\tau_1, \dots, \tau_J)$ is estimated as the average of $J$ \emph{independent} reward models $\phi_j(\tilde{\tau})$. This estimate incurs an approximation error due to unmodeled correlations between segments, which is bounded by a small constant $\varepsilon$. 

The prompt-segment independence assumption is motivated by the modular structure of prompts in PDT. Prompt segments correspond to localized behaviors, and many MDPs can be effectively identified by a small number of informative state-action pairs. In such cases, inter-segment interactions are minimal, since the presence of a few discriminative $(\hat{r}, \mathbf{s}, \mathbf{a})$ transitions is sufficient for task identification. Moreover, during PDT pretraining, prompt segments are sampled independently from the demonstration pool without constraints on order or co-occurrence. As a result, the model is not encouraged to rely on global interactions between segments, but instead learns to attend to individually informative transitions within each segment. This training setup implicitly promotes invariance to inter-segment dependencies and limits the influence of global prompt structure on downstream behavior. See Section~\ref{app:segment_attention} for an attention weight analysis that further supports this interpretation.

\begin{theorem}\label{thm:regret_bound}
Assume that the reward function $G\colon P^J \to \mathbb{R}$ for a prompt $\rho = (\tilde{\tau}_1,\dots,\tilde{\tau}_J)$ decomposes as the mean of $J$ independent reward models $\phi_j(\tilde{\tau}_j)$
\begin{equation}\label{eq:decomposition}
    G(\rho) = \frac{1}{J} \sum_{j=1}^{J} \phi_j(\tilde{\tau}_j) + h(\tilde{\tau}_1,\dots,\tilde{\tau}_J),
\end{equation}
and that the interaction term $h$ is uniformly bounded by
$
    |h(\tilde{\tau}_1,\dots,\tilde{\tau}_J)| \leq \varepsilon,\quad \forall\, \tilde{\tau}_j \in P.
$
Let $\rho^* = (\tilde{\tau}^*_1,\dots,\tilde{\tau}^*_J)$ denote the optimal prompt, and suppose that for each slot $j$, a bandit algorithm guarantees a slot-specific regret
\begin{align}
\mathrm{Regret}_j(K) = \sum_{t=1}^{K} \mathbb{E}\left[\phi_j(\tilde{\tau}^*_j)-\phi_j(\tilde{\tau}_{t,j})\right]
\end{align}
over $K$ rounds. Then the cumulative regret after $K$ rounds is bounded as:
\begin{align}
    \mathrm{Regret}(K) &\triangleq \sum_{t=1}^{K} \mathbb{E}\left[G(\rho^*) - G(\rho_t)\right] %
    \leq \frac{1}{J} \sum_{j=1}^{J} \mathrm{Regret}_j(K) + 2K\varepsilon. \label{eq:regret_bound}
\end{align}
\end{theorem}
\begin{corollary}\label{cor:regret_bound}
Under the same assumptions as Theorem~\ref{thm:regret_bound}, suppose each slot-specific reward model $\phi_j$ is learned using a standard contextual bandit algorithm (e.g., UCB, Thompson Sampling, $\epsilon$-greedy) over the set of segments $P$, with $|P|$ segments. Without loss of generality, assume that each $\phi_j$ is bounded in $[0,1]$ either via normalization of returns or bounded regression targets. If the regret for each $\phi_j$ satisfies $\mathrm{Regret}_j(K) = \mathcal{O}\left( \sqrt{K \log |P|} \right)$, then the total regret is bounded by
\begin{equation}
\mathrm{Regret}(K) = \mathcal{O}\left( \sqrt{K \log |P|} + K\varepsilon \right).    
\end{equation}

\end{corollary}
Corollary~\ref{cor:regret_bound} shows that the proposed bandit architecture preserves the sublinear regret bound of standard algorithms, while introducing an additional error term that grows linearly with the number of rounds $K$.

\subsection{Arm Features}\label{sec:arm-features}
Learning reward models $\phi_j: \Tilde{\tau} \to \mathbb{R}$ for unencoded trajectory segments becomes impractical for MDPs with large state spaces, such as those involving pixel-space observations and high-dimensional actions. The key issue lies in the input size of these reward models, which scales as $H \times (|\mathcal{S}| + |\mathcal{A}| + 1)$, growing linearly with $\mathcal{S}, \mathcal{A}$, and segment length $H$.
This limitation can be addressed by integrating a feature extractor $\Psi: {\Tilde{\tau}} \to \mathbb{R}^d$ that embeds unencoded trajectory segments from the MDP modalities into a latent feature space. 

We propose to leverage the pre-trained PDT model as a feature extractor by taking the latent representation of prompt tokens as the embedding for the prompt. This approach not only mitigates the scaling issue but also aligns with the inductive biases of the pre-trained PDT, which is expected to encode meaningful, task-relevant information. The reward models then operate on these fixed-size embeddings rather than unencoded, flattened segments, significantly improving scalability.

In summary, we introduce a specialized bandit architecture for PDT prompt tuning. By leveraging the structure of the prompt space, we optimize $J$ independent reward models, each corresponding to a segment position in the trajectory prompt, which reduces the search complexity from combinatorial to linear. Additionally, we utilize the pre-trained Transformer backbone to embed prompt segments, allowing the reward model input size to remain fixed even in high-dimensional state and action spaces.
\section{Experiments}
\label{sec:experiments}
We now present a comprehensive evaluation of our method. We first assess performance across multiple MuJoCo~\citep{todorov2012mujoco} tasks from standard multi-task benchmarks~\citep{yu2020meta, finn2017model}. We then analyze prompt-space exploration in a custom 2D environment. Finally, we evaluate the regret behavior of our bandit architecture and compare its scalability to standard algorithms.

\subsection{Environments, Datasets, Baselines}
\textbf{Environments}.
\begin{wrapfigure}[16]{r}{0.45\textwidth}
    \centering
    \begin{minipage}[t]{0.22\textwidth}
        \includegraphics[width=\linewidth]{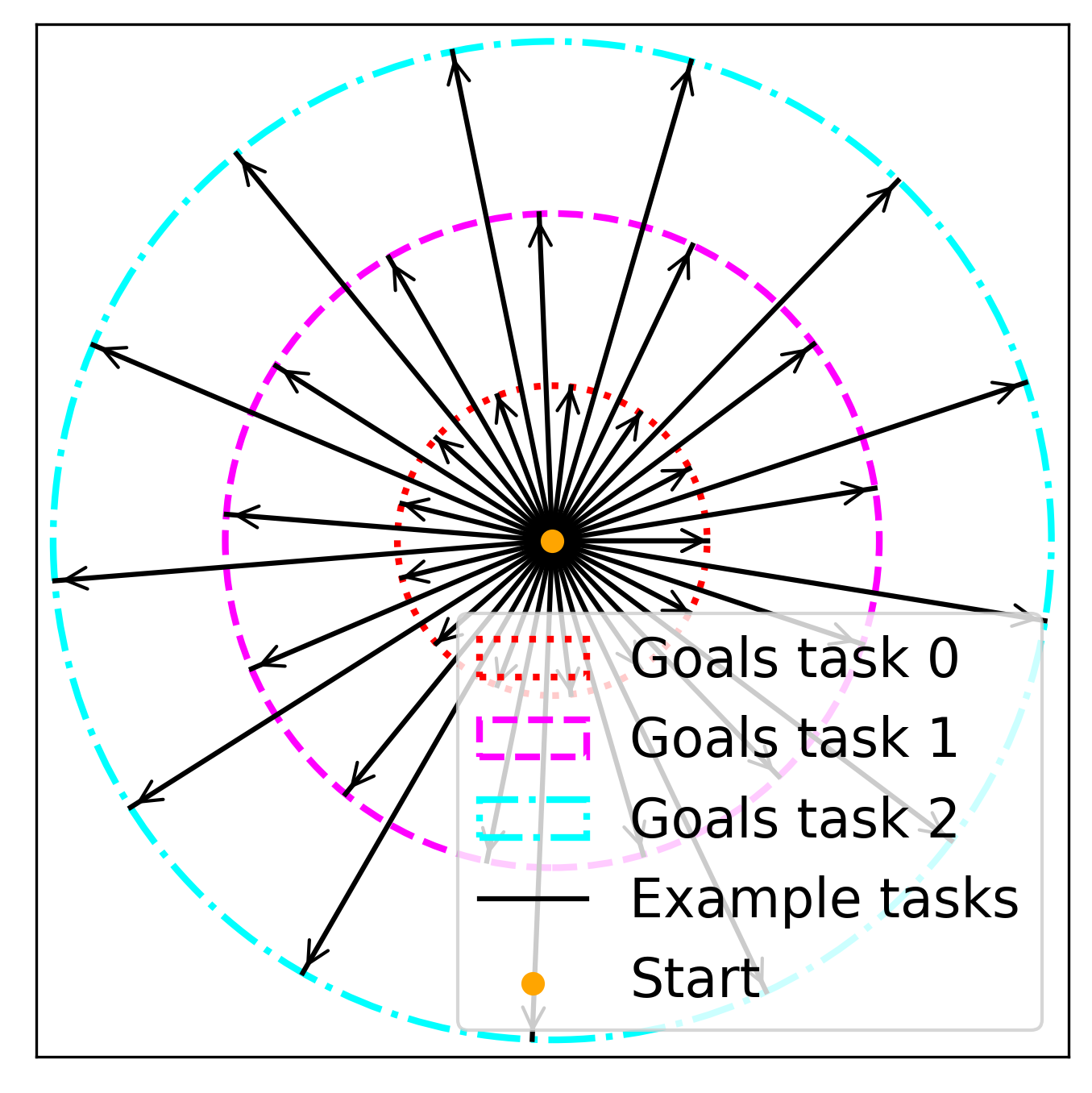}
        \subcaption{Multiple tasks.}
        \label{subfig:2d_env_tasks}
    \end{minipage}
    \hfill
    \begin{minipage}[t]{0.22\textwidth}
        \includegraphics[width=\linewidth]{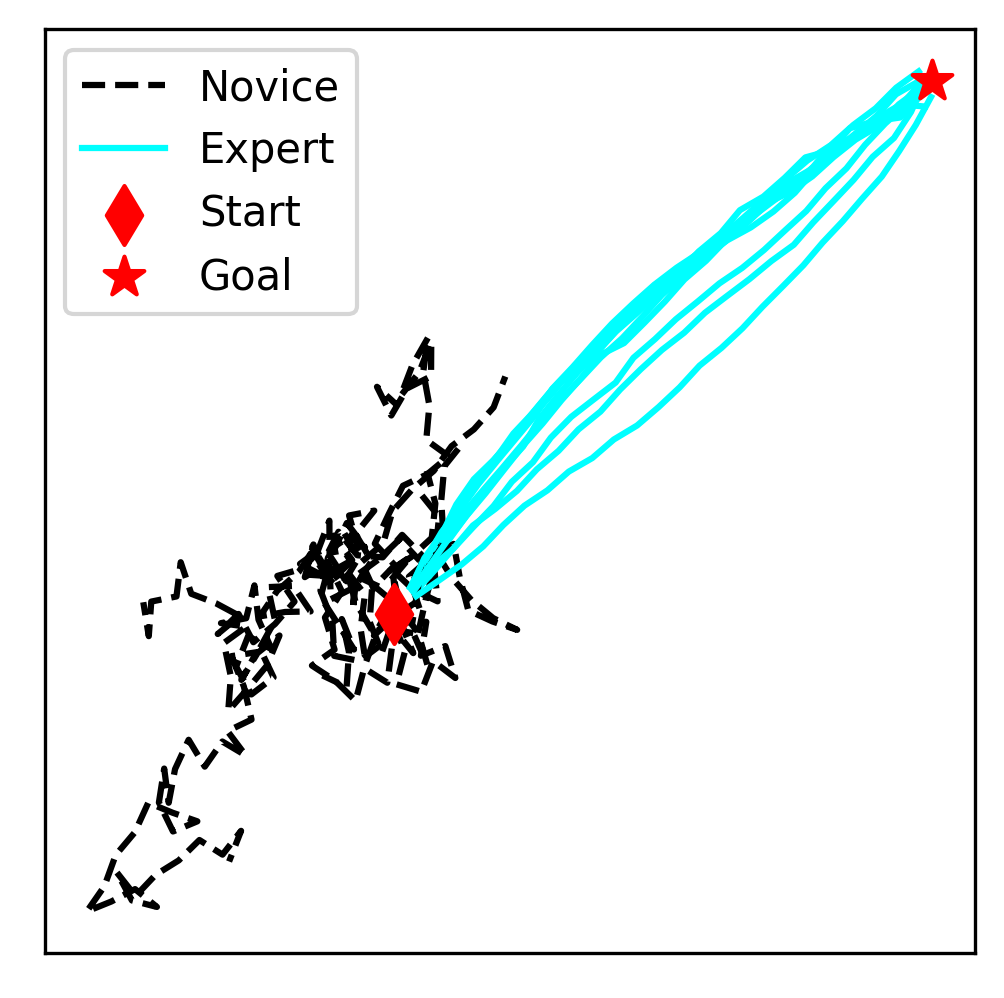}
        \subcaption{Trajectories.}
        \label{subfig:expert_vs_novice}
    \end{minipage}
    \caption{\texttt{Sparse 2D point} environment: (a) Multiple tasks via parameters $r$ (radius) and $\alpha$ (angle); (b) Expert vs. novice trajectories.}
    \label{fig:2d_env_overview}
\end{wrapfigure}

We consider the following environments: (i) \texttt{MuJoCo Half Cheetah}: Joint control of a half-cheetah agent. Tasks involve matching varying target velocities. (ii) \texttt{MuJoCo Ant}: Joint control of an ant agent. Tasks involve movement along different target directions. 
(iii) \texttt{Meta-world Pick-place}: End-effector control of a simulated Sawyer arm. Tasks require picking objects from varying locations and placing them at distinct target positions. (iv) \texttt{Sparse 2D Point}: A planar point agent navigates to a goal position. Tasks are distinguished by varying goal coordinates, distributed on circles with different radii and angles, as depicted in Figure~\ref{subfig:2d_env_tasks}. The agent must issue a \texttt{stop} action to end the episode and receive a sparse reward, defined as the negative distance to the goal. A bonus of +10 is awarded for stopping near the target. Crucially, for all settings, the task parameter (target velocity, direction, goal position) is \textit{not} part of the state -- it must be inferred from the trajectory prompt. A detailed descriptions of the environments is provided in Appendix~\ref{app:2D-env-task-split}.

\textbf{Offline RL dataset.} For each training task $i$, PDT requires a dataset of trajectories $\mathcal{T}_i$ and a dataset of (expert) demonstrations $\mathcal{P}_i$. For the \texttt{MuJoCo} tasks, we use datasets and corresponding training and testing tasks from \citet{xu2022prompting}.
To generate datasets for the \texttt{Sparse 2D point} environment, we employ PPO~\citep{schulman2017proximal} on 60 tasks with three discrete radii and 20 discrete angles. The first 48 tasks are used for training, and the remaining 12 are reserved for testing generalization. 
PPO is run for 1M steps on each training task, and the resulting trajectories are stored as $\mathcal{D}_i$. The highest-return trajectories are selected to form $\mathcal{P}_i$. To reduce the computational cost of computing $\arg\max$ over all possible prompt segments, $\mathcal{P}_i$ is limited to 10 randomly sampled high-return trajectories.

\textbf{Baselines}. We evaluate our method against the following RL baselines: (i) an optimal policy oracle, either scripted or learned using Conservative Q-Learning (CQL)~\citep{kumar2020conservative}; (ii) a Decision Transformer (DT) trained across multiple tasks without further enhancements; (iii) a vanilla Prompt Decision Transformer (PDT) that samples prompts uniformly at random; (iv) a Gaussian perturbation-based method that performs hill climbing in the prompt space; and (v) the ranking-based prompt tuner from~\cite{hu2023prompt}, referred to as ``ZORankSGD''.

\subsection{Prompt Tuning Results and Analysis}
\textbf{Does the proposed method reliably improve the performance of a frozen PDT backbone?}
We investigate this question empirically on three MuJoCo environments, where we compare the inference-time performance of a standard PDT, that samples prompts randomly from $\mathcal{P}_i$, to a PDT enhanced by our bandit-based prompt-tuning method.
Each method is evaluated over 250 online rollouts on in-distribution training tasks, and we report the mean return from the final 10 rollouts, averaged over three seeds and all tasks. Results are presented in Table~\ref{tab:inference-performance-comparison}.

\begin{table}[t]
\centering
\resizebox{\textwidth}{!}{%
\begin{tabular}{lcc|cc|cc}
\multirow{2}{*}{Method} & \multicolumn{2}{c|}{\texttt{MuJoCo Half Cheetah}} & \multicolumn{2}{c|}{\texttt{MuJoCo Ant}} & \multicolumn{2}{c}{\texttt{Meta-world Pick-place}} \\
\cline{2-7}
 & $J = 1, H = 5$ & $J = 2, H = 20$ & $J = 1, H = 5$ & $J = 2, H = 20$ & $J = 1, H = 5$ & $J = 2, H = 2$ \\
\hline
CQL oracle & -25.82 $\pm$ 12.53 & -25.82 $\pm$ 12.53 & 760.54 $\pm$ 288.20 & 760.54 $\pm$ 288.20 &  535.84 $\pm$ 31.02 &  535.84 $\pm$ 31.02  \\
PDT, no tuning & -44.75 $\pm$ 0.91 & -42.68 $\pm$ 1.3 & 694.43 $\pm$ 227.04 & 754.22 $\pm$ 175.1 &  551.58 $\pm$ 26.09 &  535.52 $\pm$ 24.86  \\
\hline
Hill-climbing & -40.00 $\pm$ 26.39 & -29.93 $\pm$ 22.84 & 738.56 $\pm$ 181.56 & 740.17 $\pm$ 189.38 &  555.79 $\pm$ 22.72  &  540.15 $\pm$ 23.49  \\
ZORankSGD & -43.62 $\pm$ 21.14 & -34.77 $\pm$ 37.70 & 735.47 $\pm$ 180.20 & 731.22 $\pm$ 172.48 &  554.26 $\pm$ 23.00  &  537.20 $\pm$ 25.5  \\
\hline
$\epsilon$-greedy$^{\Psi}$ & {-42.60 $\pm$ 3.77} & {-34.12 $\pm$ 3.12} & 815.17 $\pm$ 182.28 & \textbf{819.52 $\pm$ 191.39} &  555.35 $\pm$ 24.15  &  541.32 $\pm$ 22.91  \\
TS$^{\Psi}$  & -38.52 $\pm$ 11.21 & -27.62 $\pm$ 12.90 & 800.95 $\pm$ 184.75 & 791.06 $\pm$ 192.75 &  \textbf{556.87 $\pm$ 24.11}  &  540.82 $\pm$ 22.83 \\
UCB$^{\Psi}$  & -36.55 $\pm$ 11.93 & -26.87 $\pm$ 14.10 & 751.35 $\pm$ 192.25 & 744.55 $\pm$ 168.50 &  552.68 $\pm$ 24.77   &  538.84 $\pm$ 23.14  \\
$\epsilon$-greedy & -76.50 $\pm$ 6.68 & -58.76 $\pm$ 8.21 & 812.14 $\pm$ 176.88 & 816.77 $\pm$ 268.42 &  556.22 $\pm$ 25.16  &  \textbf{541.80 $\pm$ 23.76}  \\
TS & \textbf{-33.56 $\pm$ 13.48} &  \textbf{-26.28 $\pm$ 10.14} & \textbf{835.38 $\pm$ 171.25} & 729.65 $\pm$ 175.41 &  556.11 $\pm$ 24.56 & 541.33 $\pm$ 22.79 \\
UCB & -35.72 $\pm$ 14.96 & -26.57 $\pm$ 10.79 & 732.22 $\pm$ 180.25 & 777.29 $\pm$ 180.77  &  554.92 $\pm$ 26.06   &  538.50 $\pm$ 23.64  \\
\hline
\end{tabular}
}
\caption{Inference time performance (mean $\pm$ std) across environments with varying trajectory segments ($J$) and lengths ($H$). Bold values indicate best performance.}
\label{tab:inference-performance-comparison}
\end{table}

Our bandit-based prompt-tuning method consistently and substantially improves the performance of the frozen PDT backbone, even surpassing the single-task CQL oracle on the \texttt{MuJoCo Ant} environment. While PDT performance generally increases with larger prompt sizes (i.e., higher $J$ and $H$), our method yields even greater gains, demonstrating the clear advantage of optimized prompt selection over uniform sampling from $\mathcal{P}_i$. 
When using Transformer-encoded trajectory segments (denoted by $\Psi$), our method achieves comparable performance in terms of return to the unencoded variants, demonstrating that the PDT provides a robust and compact representation for bandit-based prompt tuning. The main benefit of this encoding lies in its fixed-dimensional representation, which makes the approach scalable to high-dimensional settings, as further evidenced by the inference-time results in Appendix~\ref{tab:wall-clock-times}.

Although hill-climbing and ranking-based prompt-tuning~\citep{hu2023prompt} can also improve performance, their results are less consistent and typically weaker. We attribute this instability to their reliance on iterative, noise-driven perturbations of initial prompts.

In summary, bandit-based prompt-tuning reliably enhances frozen PDT performance, outperforming existing baselines, particularly when leveraging Transformer-encoded prompt segments.

\textbf{Does the bandit-based method improve inference-time performance on OOD tasks?} To assess the benefits of prompt tuning in out-of-distribution (OOD) scenarios, we evaluate the frozen PDT model on OOD tasks from the MuJoCo suite using the same procedure as before. Results are shown in Table~\ref{tab:results-ood}. 
We find that our bandit-based method significantly improves PDT's generalization in all tasks, often reaching performance levels comparable to in-distribution tasks (\texttt{MuJoCo Half Cheetah}, \texttt{Meta-world Pick-Place}), though not always achieving optimal returns. 
This suggests that the effectiveness of our method extends to OOD tasks, provided the underlying model retains sufficient generalization capability.

In contrast, hill-climbing and ranking-based prompt-tuning~\citep{hu2024prompt} exhibit reduced robustness in the OOD setting, 
falling considerably short of the performance gains of our method in \texttt{MuJoCo Half Cheetah} and \texttt{MuJoCo Ant} environments. We hypothesize that this degradation arises from their reliance on the initial sampled prompt and the increased optimization difficulty faced by the PDT in unfamiliar environments.

Finally, we observe that fine-tuning the PDT backbone solely on target task data for 250 epochs does not reliable improve performance and even degrades performance in some settings. 
This degradation is hypothesized to stem from the large size of the Transformer backbone, the absence of regularization provided by diverse multi-task training data, 
and less efficient full-model updates. A similar phenomenon was previously reported by \citet{yuan2024pre}.

\begin{table}[t]
\centering
\resizebox{\textwidth}{!}{%
\begin{tabular}{lcc|cc|cc}
\multirow{2}{*}{Method} & \multicolumn{2}{c|}{\texttt{MuJoCo Half Cheetah}} & \multicolumn{2}{c|}{\texttt{MuJoCo Ant}} & \multicolumn{2}{c}{\texttt{Meta-world Pick-Place} } \\
\cline{2-7}
 & $J = 1, H = 5$ & $J = 2, H = 20$ & $J = 1, H = 5$ & $J = 2, H = 20$ & $J = 1, H = 5$ & $J = 2, H = 2$ \\
\hline
CQL Oracle & -23.8 $\pm$ 10.39 & -23.81 $\pm$ 10.39 & 508.09 $\pm$ 231.90 & 508.09 $\pm$ 231.90 & 525.07 $\pm$ 60.15 &  525.07 $\pm$ 60.15 \\
PDT, no tuning & -64.78 $\pm$ 36.91 & -40.95 $\pm$ 43.19 & 363.90 $\pm$ 105.42 & 360.07 $\pm$ 72.36 & 502.8 $\pm$ 63.98  &  524.37 $\pm$ 39.56 \\
\hline
PDT, finetuned & -129.05 $\pm$ 65.91 & -39.30 $\pm$ 14.66 & 306.29 $\pm$ 63.91 & 360.96 $\pm$ 138.58 & 495.37 $\pm$ 57.87  & 488.17 $\pm$ 50.15  \\
Hill-climbing & -53.84 $\pm$ 23.56 & -34.39 $\pm$ 26.05 & 355.80 $\pm$ 135.17 & 344.46 $\pm$ 57.21 & \textbf{560.92 $\pm$ 27.04} & 544.19 $\pm$ 28.83 \\
ZORankSGD & -59.85 $\pm$ 32.37 & -36.6 $\pm$ 19.45 & 383.57 $\pm$ 193.35 & 340.68 $\pm$ 44.60 & 503.56 $\pm$ 66.16 & 538.05 $\pm$ 31.46 \\
\hline
$\epsilon$-greedy$^{\Psi}$ & {-32.61 $\pm$ 19.85} & \textbf{-23.93 $\pm$ 14.14} & {477.24 $\pm$ 84.64} & {431.52 $\pm$ 43.69} & 531.49 $\pm$ 49.86 & 552.11 $\pm$ 23.83 \\
TS$^{\Psi}$  & -38.56 $\pm$ 21.61 & -31.80 $\pm$ 14.20 & 468.76 $\pm$ 79.50 & \textbf{441.44 $\pm$ 80.25} & 549.38 $\pm$ 36.12  & 553.12 $\pm$ 20.81 \\
UCB$^{\Psi}$  & -42.76 $\pm$ 15.77 & -33.27 $\pm$ 17.27 & 392.39 $\pm$ 51.57 & 346.59 $\pm$ 49.50 & 506.43 $\pm$ 65.38 & 539.42 $\pm$ 31.12 \\
$\epsilon$-greedy & -29.97 $\pm$ 21.10 & -25.18 $\pm$ 22.17 & \textbf{480.85 $\pm$ 84.76} & 431.74 $\pm$ 44.20 & 530.26 $\pm$ 51.61 & 550.06 $\pm$ 22.38 \\
TS  & \textbf{-29.26 $\pm$ 21.25} & -34.23 $\pm$ 14.51  & 466.11 $\pm$ 79.50 & {438.82 $\pm$ 125.40} & 551.33 $\pm$ 34.52 & \textbf{553.34 $\pm$ 18.82} \\
UCB & -42.99 $\pm$ 15.67 & -35.78 $\pm$ 16.83 & 376.68 $\pm$ 82.25 & 413.31$\pm$ 46.31 & 512.25 $\pm$ 69.62 & 534.58 $\pm$ 35.35\\
\hline
\end{tabular}
}
\caption{OOD performance (mean $\pm$ std) across environments with varying trajectory segments ($J$) and lengths ($H$). Bold values indicate best performance.}
\label{tab:results-ood}
\end{table}

\begin{figure*}
    \centering
    \begin{subfigure}[t]{0.47\textwidth}
        \centering
        \includegraphics[width=\linewidth]{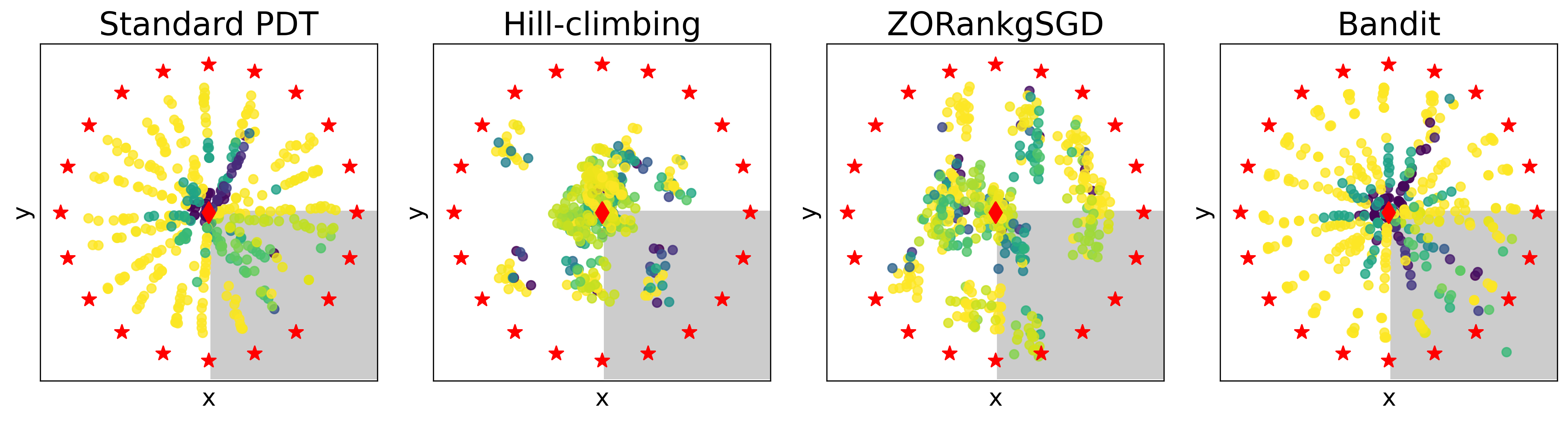}
        \includegraphics[width=\linewidth]{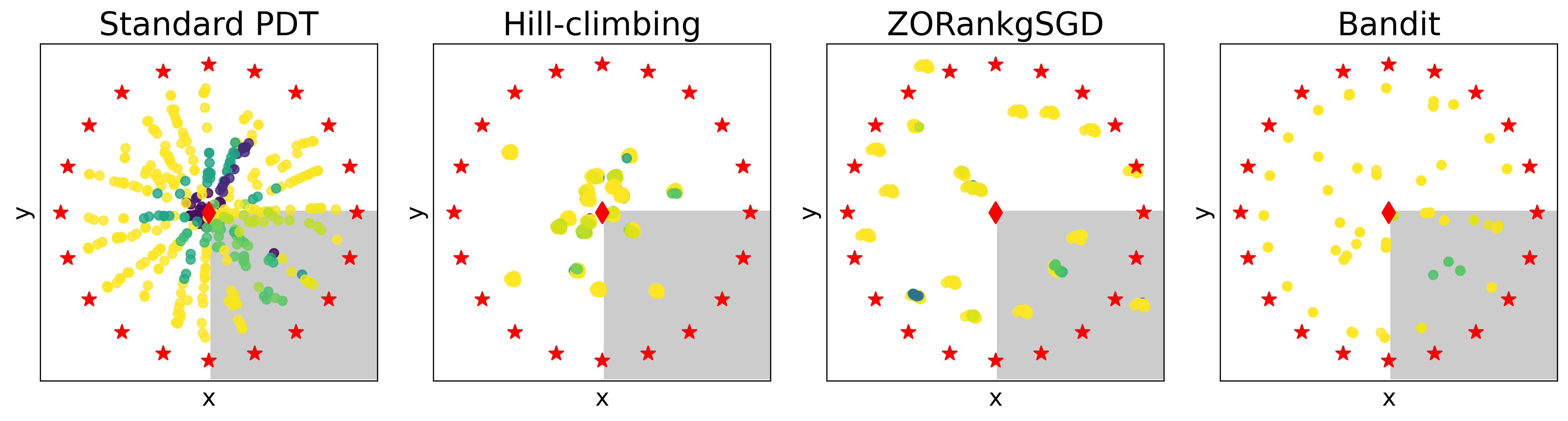}
        \caption{\textbf{Top}: First 20 prompts. \textbf{Bottom}: Last 20 prompts.
        Each dot represents the prompt's mean state coordinate, colored according to the achieved return \includegraphics[height=\baselineskip]{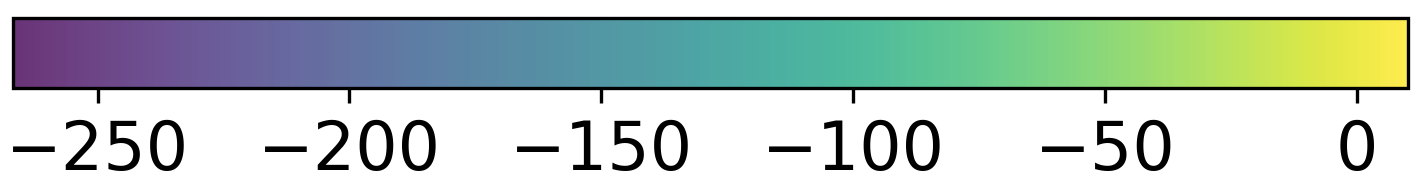}. Stars indicate task goals, the diamond marks the initial state, the shaded region indicates OOD test tasks.}
        \label{subfig:spatio-temporal}
    \end{subfigure}\hfill%
    \begin{subfigure}[t]{0.47\textwidth}
        \centering
        \resizebox{\textwidth}{!}{%
        \begin{tabular}{lccc}
            Method & $J=1$ & $J=2$ & $J=4$ \\
            \hline
            $\pi^*$ oracle & $10 \pm 0.0$ & $10 \pm 0.0$ &  $10 \pm 0.0$ \\
            PDT, no tuning & 0.0 $\pm$ 2.1 & 6.3 $\pm$ 0.8 & 8.3 $\pm$ 0.6 \\
            \hline
            Hill-climbing & 5.8 $\pm$ 3.8 & 7.9 $\pm$ 1.6 & 6.2 $\pm$ 4.0 \\
            ZORankSGD & -0.6 $\pm$ 30.9 & 4.4 $\pm$ 16.7 & 3.1 $\pm$ 22.1 \\
            \hline
            $\epsilon$-greedy$^\Psi$ & 9.0 $\pm$ 0.6 & 9.4 $\pm$ 0.3 & {9.6 $\pm$ 0.2} \\
            UCB$^\Psi$ & 9.2 $\pm$ 0.5 & 8.8 $\pm$ 0.9 & 8.8 $\pm$ 0.9 \\
            TS$^\Psi$ & 9.7 $\pm$ 0.1 & 9.7 $\pm$ 0.2 & 9.4 $\pm$ 0.5 \\
            $\epsilon$-greedy & 8.9 $\pm$ 0.5 & {9.5 $\pm$ 0.3} & 8.9 $\pm$ 0.7 \\
            UCB & {9.4 $\pm$ 0.5} & {9.5 $\pm$ 0.3} & 9.3 $\pm$ 0.4 \\
            TS & \textbf{9.9 $\pm$ 0.0} & \textbf{9.9 $\pm$ 0.0} & \textbf{9.8 $\pm$ 0.1} \\
        \end{tabular}
        }
        \caption{Inference-time performance (mean $\pm$ std) in the \texttt{Sparse 2D Point} environment. A standard single-task Decision DT achieves $-64.3 \pm 24.1$ average return, underscoring the need for multi-task models like PDT. Results are averaged over training tasks, three seeds, and the final 50 rollouts.}
        \label{tab:2D-id-results}
    \end{subfigure}
    \caption{(a) Visualization of prompt selection across tasks in the \texttt{Sparse 2D Point} environment. (b) Inference-time performance showing the benefit of prompt tuning over a single-task baseline.}
    \label{fig:spatio-temporal}
\end{figure*}

\textbf{How do different methods explore the prompt space?} We use the \texttt{Sparse 2D Point} environment to qualitatively analyze how different methods explore the prompt space. For interpretability, we consider a PDT with a single prompt segment ($J = 1$), allowing prompts to be visualized via the mean spatial coordinate of their states. Each method is run for 250 rollouts per task, and we visualize the first and last 20 selected prompts in Figure~\ref{subfig:spatio-temporal}.

The standard PDT samples the prompt space uniformly without refining its selection strategy over time. Hill-climbing gradually discovers high-performing prompts in the local neighborhood of the initial sample by iteratively applying Gaussian noise. ZORankSGD moves prompts toward the task goal, sometimes overshooting it and generating out-of-distribution prompts due to the unconstrained generative approach of~\citet{hu2023prompt}.

Our bandit method (with $\epsilon$-greedy exploration) initially samples prompts uniformly, capturing both low- and high-performing examples. Over time, it leverages accumulated reward estimates to avoid low-reward regions and increasingly selects segments near the task goal, which provide more informative cues for task identification.

We also report the performance of all methods on the 2D environment tasks with the largest radius in Table~\ref{tab:2D-id-results}. 
The standard PDT without prompt tuning falls drastically short of the optimal return, due to the overlap between prompt datasets. The performance of the PDT increases with larger prompts, which implies that excessive random sampling suffices to find informative prompts in this environment.
Best prompt-tuning results are again achieved by our bandits compared to the hill-climbing and ranking-based prompt-tuning baselines, with Thompson Sampling (TS) performing slightly better than UCB or $\epsilon$-greedy exploration. We find no considerable difference between using Transformer encoded prompts (marked by $\Psi$) or not, which can be attributed to the simplicity of the 2D environment.

\subsection{Bandit Regret Analysis}\label{sec:regret_experiment}
We empirically evaluate the proposed bandit-based method with respect to the interaction bound $\varepsilon$, and compare its performance to standard MAB algorithms that learn directly over the full combinatorial space. This evaluation is conducted on a synthetic task carefully designed to reflect key characteristics of PDT prompt tuning. Specifically, the task involves identifying an optimal vector $\mathbf{x}$ of length $J$, where each dimension takes values in $\{1, \dots, H\}$. The reward function $R(\mathbf{x})$ is defined only for complete vectors and is bounded within $[0, 1]$, making the contribution of individual components unobservable, mirroring the structure of real-world setting.
\begin{figure}
    \centering
    \includegraphics[width=\linewidth]{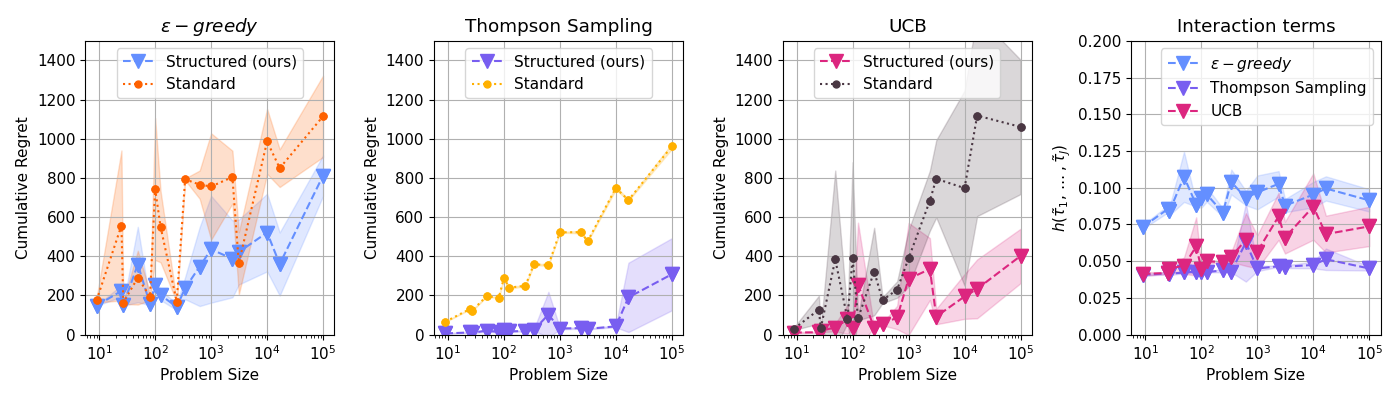}
    \caption{Performance of the structured (ours) and standard MAB methods on a synthetic prompt tuning task. Problem instances are generated by sweeping over $J = \{2, 3, 4, 5\}$ segments and $H = \{3, 5, 7, 10\}$ choices, with problem size reported as $H^J$. Shaded regions indicate one standard deviation around the mean; results are averaged over three random seeds.}
    \label{fig:synthetic_bandit_results}
\end{figure}
While the synthetic setup is necessarily simplified, it enables controlled investigation of core assumptions underlying our theoretical analysis, particularly the independence of segment-wise reward estimates. We report cumulative regret after $K = 2000$ rounds for varying values of $J$ and $H$, comparing our structured method to standard $\epsilon$-greedy, UCB, and TS strategies. For our method, we also estimate the neglected interaction term $h(\mathbf{x}_1, \dots, \mathbf{x}_J) = \left| R(\mathbf{x}) - \frac{1}{J} \sum_j \phi_j(\mathbf{x}_j) \right|$, where $R(\mathbf{x})$ denotes the true reward.

As shown in Figure~\ref{fig:synthetic_bandit_results}, our method matches the performance of standard, ``flat'' bandit baselines on small problems and consistently outperforms them as problem size grows. This demonstrates more efficient exploration of the combinatorial space by leveraging its structure. Additionally, the interaction term $h(\mathbf{x}_1, \dots, \mathbf{x}_J)$ remains low and increases only marginally with problem size, empirically supporting the bounded interaction assumption in Theorem~\ref{thm:regret_bound}. These findings reinforce the validity of our method even in settings where interactions between prompt segments exists while global reward must be inferred from segment-level estimates.

\section{Conclusion}
\label{sec:conclusion}
This work introduces a bandit-based prompt-tuning method extending PDT and addressing the limitations of uniform prompt sampling. By leveraging a contextual MAB framework, our approach optimizes the selection of trajectory segments to maximize task performance and enhance adaptation to unseen tasks. Experimental results highlight consistent improvements across diverse tasks, demonstrating the efficacy and robustness of the proposed method in multiple environments. 
This method not only advances the state of prompt optimization in PDT but also contributes to the broader integration of offline RL and sequence modeling paradigms.

\textbf{Limitation and broader impact.} 
A key limitation is the combinatorial expansion of the search space as the number of demonstrations increases, which renders online exploration impractical. A promising direction is to learn a sampler that pre-selects high-potential segments or clusters similar prompts to reduce redundancy. While bandit-based tuning enhances OOD performance, pure offline RL struggles with extrapolation. Incorporating meta-learning techniques, such as In-Context RL~\citep{laskin2022context}, into PDT pretraining offers a compelling extension of this work.

\section*{Acknowledgments}
This work was partially funded by Wallenberg Autonomous Systems Programs (WASP).

\bibliography{main}
\bibliographystyle{iclr2025_conference}

\newpage

\setcounter{page}{1}

\section*{NeurIPS Paper Checklist}

\begin{enumerate}

\item {\bf Claims}
    \item[] Question: Do the main claims made in the abstract and introduction accurately reflect the paper's contributions and scope?
    \item[] Answer: \answerYes{} %
    \item[] Justification: The contributions outlined in the abstract and introduction are consistent with the prompt-tuning bandit method Section~\ref{sec:method} the evaluations in Section~\ref{sec:experiments}.
    \item[] Guidelines:
    \begin{itemize}
        \item The answer NA means that the abstract and introduction do not include the claims made in the paper.
        \item The abstract and/or introduction should clearly state the claims made, including the contributions made in the paper and important assumptions and limitations. A No or NA answer to this question will not be perceived well by the reviewers. 
        \item The claims made should match theoretical and experimental results, and reflect how much the results can be expected to generalize to other settings. 
        \item It is fine to include aspirational goals as motivation as long as it is clear that these goals are not attained by the paper. 
    \end{itemize}

\item {\bf Limitations}
    \item[] Question: Does the paper discuss the limitations of the work performed by the authors?
    \item[] Answer: \answerYes{} %
    \item[] Justification: We discuss the main limitations of our method, and potential remedies, in Section~\ref{sec:conclusion}. Potential limitations in the evaluation procedure are discussed in Section~\ref{sec:experiments}.
    \item[] Guidelines:
    \begin{itemize}
        \item The answer NA means that the paper has no limitation while the answer No means that the paper has limitations, but those are not discussed in the paper. 
        \item The authors are encouraged to create a separate "Limitations" section in their paper.
        \item The paper should point out any strong assumptions and how robust the results are to violations of these assumptions (e.g., independence assumptions, noiseless settings, model well-specification, asymptotic approximations only holding locally). The authors should reflect on how these assumptions might be violated in practice and what the implications would be.
        \item The authors should reflect on the scope of the claims made, e.g., if the approach was only tested on a few datasets or with a few runs. In general, empirical results often depend on implicit assumptions, which should be articulated.
        \item The authors should reflect on the factors that influence the performance of the approach. For example, a facial recognition algorithm may perform poorly when image resolution is low or images are taken in low lighting. Or a speech-to-text system might not be used reliably to provide closed captions for online lectures because it fails to handle technical jargon.
        \item The authors should discuss the computational efficiency of the proposed algorithms and how they scale with dataset size.
        \item If applicable, the authors should discuss possible limitations of their approach to address problems of privacy and fairness.
        \item While the authors might fear that complete honesty about limitations might be used by reviewers as grounds for rejection, a worse outcome might be that reviewers discover limitations that aren't acknowledged in the paper. The authors should use their best judgment and recognize that individual actions in favor of transparency play an important role in developing norms that preserve the integrity of the community. Reviewers will be specifically instructed to not penalize honesty concerning limitations.
    \end{itemize}

\item {\bf Theory assumptions and proofs}
    \item[] Question: For each theoretical result, does the paper provide the full set of assumptions and a complete (and correct) proof?
    \item[] Answer: \answerYes{} %
    \item[] Justification: Our theoretical result is the regret bound of our bandit architecture in Theorem~\ref{thm:regret_bound}. We provide a proof for this regret bound in Section~\ref{appx:regret_proof} in the Appendix and state all assumptions.
    \item[] Guidelines:
    \begin{itemize}
        \item The answer NA means that the paper does not include theoretical results. 
        \item All the theorems, formulas, and proofs in the paper should be numbered and cross-referenced.
        \item All assumptions should be clearly stated or referenced in the statement of any theorems.
        \item The proofs can either appear in the main paper or the supplemental material, but if they appear in the supplemental material, the authors are encouraged to provide a short proof sketch to provide intuition. 
        \item Inversely, any informal proof provided in the core of the paper should be complemented by formal proofs provided in appendix or supplemental material.
        \item Theorems and Lemmas that the proof relies upon should be properly referenced. 
    \end{itemize}

    \item {\bf Experimental result reproducibility}
    \item[] Question: Does the paper fully disclose all the information needed to reproduce the main experimental results of the paper to the extent that it affects the main claims and/or conclusions of the paper (regardless of whether the code and data are provided or not)?
    \item[] Answer: \answerYes{} %
    \item[] Justification: We provide the complete pseudocode of our method in Section~\ref{app:algorithms} in the Appendix. We provide a detailed description of our custom \texttt{Sparse 2D point} environment in Section~\ref{app:2D-env-task-split}. Training details and hyperparameters are stated in Section~\ref{appx:training_details} in the Appendix.
    \item[] Guidelines:
    \begin{itemize}
        \item The answer NA means that the paper does not include experiments.
        \item If the paper includes experiments, a No answer to this question will not be perceived well by the reviewers: Making the paper reproducible is important, regardless of whether the code and data are provided or not.
        \item If the contribution is a dataset and/or model, the authors should describe the steps taken to make their results reproducible or verifiable. 
        \item Depending on the contribution, reproducibility can be accomplished in various ways. For example, if the contribution is a novel architecture, describing the architecture fully might suffice, or if the contribution is a specific model and empirical evaluation, it may be necessary to either make it possible for others to replicate the model with the same dataset, or provide access to the model. In general. releasing code and data is often one good way to accomplish this, but reproducibility can also be provided via detailed instructions for how to replicate the results, access to a hosted model (e.g., in the case of a large language model), releasing of a model checkpoint, or other means that are appropriate to the research performed.
        \item While NeurIPS does not require releasing code, the conference does require all submissions to provide some reasonable avenue for reproducibility, which may depend on the nature of the contribution. For example
        \begin{enumerate}
            \item If the contribution is primarily a new algorithm, the paper should make it clear how to reproduce that algorithm.
            \item If the contribution is primarily a new model architecture, the paper should describe the architecture clearly and fully.
            \item If the contribution is a new model (e.g., a large language model), then there should either be a way to access this model for reproducing the results or a way to reproduce the model (e.g., with an open-source dataset or instructions for how to construct the dataset).
            \item We recognize that reproducibility may be tricky in some cases, in which case authors are welcome to describe the particular way they provide for reproducibility. In the case of closed-source models, it may be that access to the model is limited in some way (e.g., to registered users), but it should be possible for other researchers to have some path to reproducing or verifying the results.
        \end{enumerate}
    \end{itemize}

\item {\bf Open access to data and code}
    \item[] Question: Does the paper provide open access to the data and code, with sufficient instructions to faithfully reproduce the main experimental results, as described in supplemental material?
    \item[] Answer: \answerYes{} %
    \item[] Justification: We provide a zip archive of our codebase with instructions for reproducing each experiment in the supplementary material.
    \item[] Guidelines:
    \begin{itemize}
        \item The answer NA means that paper does not include experiments requiring code.
        \item Please see the NeurIPS code and data submission guidelines (\url{https://nips.cc/public/guides/CodeSubmissionPolicy}) for more details.
        \item While we encourage the release of code and data, we understand that this might not be possible, so “No” is an acceptable answer. Papers cannot be rejected simply for not including code, unless this is central to the contribution (e.g., for a new open-source benchmark).
        \item The instructions should contain the exact command and environment needed to run to reproduce the results. See the NeurIPS code and data submission guidelines (\url{https://nips.cc/public/guides/CodeSubmissionPolicy}) for more details.
        \item The authors should provide instructions on data access and preparation, including how to access the raw data, preprocessed data, intermediate data, and generated data, etc.
        \item The authors should provide scripts to reproduce all experimental results for the new proposed method and baselines. If only a subset of experiments are reproducible, they should state which ones are omitted from the script and why.
        \item At submission time, to preserve anonymity, the authors should release anonymized versions (if applicable).
        \item Providing as much information as possible in supplemental material (appended to the paper) is recommended, but including URLs to data and code is permitted.
    \end{itemize}

\item {\bf Experimental setting/details}
    \item[] Question: Does the paper specify all the training and test details (e.g., data splits, hyperparameters, how they were chosen, type of optimizer, etc.) necessary to understand the results?
    \item[] Answer: \answerYes{} %
    \item[] Justification: We state the split between in-distribution training tasks and out-of-distribution testing tasks for each environment in Section~\ref{app:2D-env-task-split} in the Appendix. Hyperparameters are stated in Section~\ref{appx:training_details}.
    \item[] Guidelines:
    \begin{itemize}
        \item The answer NA means that the paper does not include experiments.
        \item The experimental setting should be presented in the core of the paper to a level of detail that is necessary to appreciate the results and make sense of them.
        \item The full details can be provided either with the code, in appendix, or as supplemental material.
    \end{itemize}

\item {\bf Experiment statistical significance}
    \item[] Question: Does the paper report error bars suitably and correctly defined or other appropriate information about the statistical significance of the experiments?
    \item[] Answer: \answerYes{} %
    \item[] Justification: We report all results in terms terms of the mean $\pm$ one standard deviation, either in tabular form or as confidence intervals in plots. 
    \item[] Guidelines:
    \begin{itemize}
        \item The answer NA means that the paper does not include experiments.
        \item The authors should answer "Yes" if the results are accompanied by error bars, confidence intervals, or statistical significance tests, at least for the experiments that support the main claims of the paper.
        \item The factors of variability that the error bars are capturing should be clearly stated (for example, train/test split, initialization, random drawing of some parameter, or overall run with given experimental conditions).
        \item The method for calculating the error bars should be explained (closed form formula, call to a library function, bootstrap, etc.)
        \item The assumptions made should be given (e.g., Normally distributed errors).
        \item It should be clear whether the error bar is the standard deviation or the standard error of the mean.
        \item It is OK to report 1-sigma error bars, but one should state it. The authors should preferably report a 2-sigma error bar than state that they have a 96\% CI, if the hypothesis of Normality of errors is not verified.
        \item For asymmetric distributions, the authors should be careful not to show in tables or figures symmetric error bars that would yield results that are out of range (e.g. negative error rates).
        \item If error bars are reported in tables or plots, The authors should explain in the text how they were calculated and reference the corresponding figures or tables in the text.
    \end{itemize}

\item {\bf Experiments compute resources}
    \item[] Question: For each experiment, does the paper provide sufficient information on the computer resources (type of compute workers, memory, time of execution) needed to reproduce the experiments?
    \item[] Answer: \answerYes{} %
    \item[] Justification: We report the hardware specification and approximate execution time for our experiments in Section~\ref{appx:training_details}.
    \item[] Guidelines:
    \begin{itemize}
        \item The answer NA means that the paper does not include experiments.
        \item The paper should indicate the type of compute workers CPU or GPU, internal cluster, or cloud provider, including relevant memory and storage.
        \item The paper should provide the amount of compute required for each of the individual experimental runs as well as estimate the total compute. 
        \item The paper should disclose whether the full research project required more compute than the experiments reported in the paper (e.g., preliminary or failed experiments that didn't make it into the paper). 
    \end{itemize}
    
\item {\bf Code of ethics}
    \item[] Question: Does the research conducted in the paper conform, in every respect, with the NeurIPS Code of Ethics \url{https://neurips.cc/public/EthicsGuidelines}?
    \item[] Answer: \answerYes{} %
    \item[] Justification: We have reviewed and adhere to the NeurIPS Code of Ethics. We experiment mainly on publically available meta RL datasets~\citep{yu2020meta, duan2016benchmarking}, additional experiments were conducted on a simulated 2D environment and on a synthetic bandit benchmark.
    \item[] Guidelines:
    \begin{itemize}
        \item The answer NA means that the authors have not reviewed the NeurIPS Code of Ethics.
        \item If the authors answer No, they should explain the special circumstances that require a deviation from the Code of Ethics.
        \item The authors should make sure to preserve anonymity (e.g., if there is a special consideration due to laws or regulations in their jurisdiction).
    \end{itemize}

\item {\bf Broader impacts}
    \item[] Question: Does the paper discuss both potential positive societal impacts and negative societal impacts of the work performed?
    \item[] Answer: \answerNA{} %
    \item[] Justification: We are contributing a foundational method to improve the performance of PDT models and can therefore not estimate the true societal impact. However, we don't see any application or deployment with negative societal impact resulting directly from this work.
    \item[] Guidelines:
    \begin{itemize}
        \item The answer NA means that there is no societal impact of the work performed.
        \item If the authors answer NA or No, they should explain why their work has no societal impact or why the paper does not address societal impact.
        \item Examples of negative societal impacts include potential malicious or unintended uses (e.g., disinformation, generating fake profiles, surveillance), fairness considerations (e.g., deployment of technologies that could make decisions that unfairly impact specific groups), privacy considerations, and security considerations.
        \item The conference expects that many papers will be foundational research and not tied to particular applications, let alone deployments. However, if there is a direct path to any negative applications, the authors should point it out. For example, it is legitimate to point out that an improvement in the quality of generative models could be used to generate deepfakes for disinformation. On the other hand, it is not needed to point out that a generic algorithm for optimizing neural networks could enable people to train models that generate Deepfakes faster.
        \item The authors should consider possible harms that could arise when the technology is being used as intended and functioning correctly, harms that could arise when the technology is being used as intended but gives incorrect results, and harms following from (intentional or unintentional) misuse of the technology.
        \item If there are negative societal impacts, the authors could also discuss possible mitigation strategies (e.g., gated release of models, providing defenses in addition to attacks, mechanisms for monitoring misuse, mechanisms to monitor how a system learns from feedback over time, improving the efficiency and accessibility of ML).
    \end{itemize}
    
\item {\bf Safeguards}
    \item[] Question: Does the paper describe safeguards that have been put in place for responsible release of data or models that have a high risk for misuse (e.g., pretrained language models, image generators, or scraped datasets)?
    \item[] Answer: \answerNA{} %
    \item[] Justification: This work does not release data or models with high risk of misuse.
    \item[] Guidelines:
    \begin{itemize}
        \item The answer NA means that the paper poses no such risks.
        \item Released models that have a high risk for misuse or dual-use should be released with necessary safeguards to allow for controlled use of the model, for example by requiring that users adhere to usage guidelines or restrictions to access the model or implementing safety filters. 
        \item Datasets that have been scraped from the Internet could pose safety risks. The authors should describe how they avoided releasing unsafe images.
        \item We recognize that providing effective safeguards is challenging, and many papers do not require this, but we encourage authors to take this into account and make a best faith effort.
    \end{itemize}

\item {\bf Licenses for existing assets}
    \item[] Question: Are the creators or original owners of assets (e.g., code, data, models), used in the paper, properly credited and are the license and terms of use explicitly mentioned and properly respected?
    \item[] Answer: \answerYes{} %
    \item[] Justification: All used assets (public datasets, algorithm implementations) are cited appropriately.
    \item[] Guidelines:
    \begin{itemize}
        \item The answer NA means that the paper does not use existing assets.
        \item The authors should cite the original paper that produced the code package or dataset.
        \item The authors should state which version of the asset is used and, if possible, include a URL.
        \item The name of the license (e.g., CC-BY 4.0) should be included for each asset.
        \item For scraped data from a particular source (e.g., website), the copyright and terms of service of that source should be provided.
        \item If assets are released, the license, copyright information, and terms of use in the package should be provided. For popular datasets, \url{paperswithcode.com/datasets} has curated licenses for some datasets. Their licensing guide can help determine the license of a dataset.
        \item For existing datasets that are re-packaged, both the original license and the license of the derived asset (if it has changed) should be provided.
        \item If this information is not available online, the authors are encouraged to reach out to the asset's creators.
    \end{itemize}

\item {\bf New assets}
    \item[] Question: Are new assets introduced in the paper well documented and is the documentation provided alongside the assets?
    \item[] Answer: \answerYes{} %
    \item[] Justification: The source code released along with our paper is properly documented and contains the license terms.
    \item[] Guidelines:
    \begin{itemize}
        \item The answer NA means that the paper does not release new assets.
        \item Researchers should communicate the details of the dataset/code/model as part of their submissions via structured templates. This includes details about training, license, limitations, etc. 
        \item The paper should discuss whether and how consent was obtained from people whose asset is used.
        \item At submission time, remember to anonymize your assets (if applicable). You can either create an anonymized URL or include an anonymized zip file.
    \end{itemize}

\item {\bf Crowdsourcing and research with human subjects}
    \item[] Question: For crowdsourcing experiments and research with human subjects, does the paper include the full text of instructions given to participants and screenshots, if applicable, as well as details about compensation (if any)? 
    \item[] Answer: \answerNA{} %
    \item[] Justification: This work did not involve crowdsourcing.
    \item[] Guidelines:
    \begin{itemize}
        \item The answer NA means that the paper does not involve crowdsourcing nor research with human subjects.
        \item Including this information in the supplemental material is fine, but if the main contribution of the paper involves human subjects, then as much detail as possible should be included in the main paper. 
        \item According to the NeurIPS Code of Ethics, workers involved in data collection, curation, or other labor should be paid at least the minimum wage in the country of the data collector. 
    \end{itemize}

\item {\bf Institutional review board (IRB) approvals or equivalent for research with human subjects}
    \item[] Question: Does the paper describe potential risks incurred by study participants, whether such risks were disclosed to the subjects, and whether Institutional Review Board (IRB) approvals (or an equivalent approval/review based on the requirements of your country or institution) were obtained?
    \item[] Answer: \answerNA{} %
    \item[] Justification: This work did not involve human or animal subjects.
    \item[] Guidelines:
    \begin{itemize}
        \item The answer NA means that the paper does not involve crowdsourcing nor research with human subjects.
        \item Depending on the country in which research is conducted, IRB approval (or equivalent) may be required for any human subjects research. If you obtained IRB approval, you should clearly state this in the paper. 
        \item We recognize that the procedures for this may vary significantly between institutions and locations, and we expect authors to adhere to the NeurIPS Code of Ethics and the guidelines for their institution. 
        \item For initial submissions, do not include any information that would break anonymity (if applicable), such as the institution conducting the review.
    \end{itemize}

\item {\bf Declaration of LLM usage}
    \item[] Question: Does the paper describe the usage of LLMs if it is an important, original, or non-standard component of the core methods in this research? Note that if the LLM is used only for writing, editing, or formatting purposes and does not impact the core methodology, scientific rigorousness, or originality of the research, declaration is not required.
    \item[] Answer: \answerNA{} %
    \item[] Justification: This work does not involve any LLM components.
    \item[] Guidelines:
    \begin{itemize}
        \item The answer NA means that the core method development in this research does not involve LLMs as any important, original, or non-standard components.
        \item Please refer to our LLM policy (\url{https://neurips.cc/Conferences/2025/LLM}) for what should or should not be described.
    \end{itemize}

\end{enumerate}

\clearpage

\appendix

\vbox{%
\hsize\textwidth
\linewidth\hsize
\vskip 0.1in
\centering
{\LARGE\bf Supplementary Material: \par}
\vspace{2\baselineskip}
}

\section{Prompt-Tuning Algorithm}\label{app:algorithms}
We provide the full pseudocode of our method. The overall prompt-tuning procedure is outlined in Algorithm~\ref{alg:bandit_steps}, prompt selection via the structured bandit is shown in Algorithm~\ref{alg:select_prompt}, and Algorithm~\ref{alg:update_bandit} details the update of the bandit's internal reward models.
\begin{figure}[H]
        \begin{algorithm}[H]
            \caption{Prompt-Tuning Bandit}\label{alg:bandit_steps}
            \begin{algorithmic}[1]
                \item[] \textbf{Input:} Pre-trained PDT parameter $\theta$, Simulator $\mathcal{M}_i$, expert demonstrations $\mathcal{P}_i$
                \item[] \textbf{Initialize:} Bandit parameter $\phi^0$, dataset $\mathcal{B} \leftarrow \empty \{ \}$
                \FOR{bandit steps $k \in K$}
                    \STATE \textbf{Algorithm~\ref{alg:select_prompt}}: Select prompt $\rho_k$ for current rollout $k$ using parameters $\phi^k$
                    \STATE Performance metric $G_i^k = 0$
                    \FOR{MDP steps $t \in T$}
                        \STATE Make PDT input (Equation~\ref{eq:pdt-seq}): $\mathbf{x}_k = \rho_k \odot \omega_{L:t}$
                        \STATE Sample action from PDT: $\mathbf{a}_t \sim \pi^*(\mathbf{x}_k; \theta)$
                        \STATE Step environment: $r_t, \mathbf{s}_{t+1} \sim \mathcal{M}_i (\mathbf{s}_t, \mathbf{a}_t)$
                        \STATE Log reward: $G_i^k = G_i^k + r_t$
                    \ENDFOR
                    \STATE Store data $\mathcal{B} \leftarrow B \cup \langle \rho_k, G_i^k \rangle$
                    \STATE \textbf{Algorithm~\ref{alg:update_bandit}}: Update $\phi^k$ using $\mathcal{B}$, yielding $\phi^{k+1}$
                \ENDFOR
                \item[] \textbf{Return:} Final bandit parameter $\phi^K$
            \end{algorithmic}
        \end{algorithm}
\end{figure}
\begin{figure}[H]
    \begin{minipage}[t]{0.48\textwidth} %
\begin{algorithm}[H]
            \caption{Prompt Selection}\label{alg:select_prompt}
            \begin{algorithmic}[1]
                
                \item[] \textbf{Input:} Expert demonstrations~$\mathcal{P}_i$, bandit parameter~$\phi = \langle \phi_1, \cdots, \phi_J \rangle$
                \item[] \textbf{Initialize:} Prediction matrix $\mathbf{Y} = [\ ]$
                \FOR{segment $\Tilde{\tau} \in \mathcal{P}_i$}
                \STATE Predict reward $ \mathbf{\hat{y}} = [ \phi_1(\Tilde{\tau}), \dots, \phi_J(\Tilde{\tau}) ]$
                \STATE Append row to prediction matrix $\begin{bmatrix} \mathbf{Y} \\ \mathbf{\hat{y}} \end{bmatrix}$
                \ENDFOR
                \item[] \textbf{Return:} $\rho = \mathcal{P}_i[\arg \max(\mathbf{Y}[j, :])]\ \forall j \in \{1, ..., J\}$
            \end{algorithmic}
        \end{algorithm}
    \end{minipage}%
    \hfill
    \begin{minipage}[t]{0.48\textwidth} %
        \begin{algorithm}[H]
            \caption{Bandit Update}\label{alg:update_bandit}
            \begin{algorithmic}[1]
            \item[] \textbf{Input:} Bandit dataset~$\mathcal{B} = \mathbf{X}, \mathbf{y}$, bandit parameter~$\phi = \langle \phi_1, \cdots, \phi_J \rangle$, learning rate~$\alpha$
                \FOR{reward model $j \in J$}
                    \STATE Get segments at $j$-th index $\mathbf{X}_j = \mathbf{X}_{[j]}$
                    \STATE Predict reward $\mathbf{\hat{y}}_j = \phi_j(\mathbf{X}_j)$
                    \STATE Define $\mathcal{L}(\phi_j) = \text{MSE}(\mathbf{\hat{y}}_j, \mathbf{y})$
                    \FOR{gradient steps $l = 0, \dots, L$}
                    \STATE $\phi_j^{l+1} = \phi_j^l - \alpha \nabla \mathcal{L}(\phi_j^l)$
                    \ENDFOR                        
                \ENDFOR
            \item[] \textbf{Return:} New parameter $\phi = \langle \phi_1^L, \cdots, \phi_J^L \rangle$
        \end{algorithmic}
        \end{algorithm}
    \end{minipage}
\end{figure}
\clearpage

\section{Environment Details}\label{app:2D-env-task-split}
\textbf{{Sparse 2D point}}: This environment involves mixed control of a planar point agent, where the state representation consists of the agent's current 2D coordinates. The agent always begins at $(0,0)$ with the objective of reaching a (hidden) goal coordinate. It has two continuous actions for movement across the plane and a binary \texttt{stop} action. The task requires the agent to navigate to the goal coordinate and appropriately select the \texttt{stop} action. A sparse reward is provided upon selecting the \texttt{stop} action, proportional to the agent's distance from the goal. When \texttt{stop} is selected within close proximity of the goal, the environment provides a reward bonus of 10, which is discounted based on the number of wasted steps. 
While the goal is not explicitly part of the state, it is implicitly encoded through the reward function, ensuring that each individual task remains a fully observable MDP.

The environment offers a continuous task space, parameterized by the angle and radius, for arbitrary goal locations on the 2D plane.
We discretize the task space by using three discrete radii, $(0.9, 1.9, 2.9)$, and 20 discrete angles $(0.0 \cdot \pi, 0.1 \cdot \pi, \dots, 1.9 \cdot \pi)$, instead of sampling continuous task parameters. 
This is primarily done to separate datasets for different tasks, since PDT requires expert demonstration for each training task. 
To separate these $3 \cdot 20 = 60$ tasks into the training set $\mathcal{T}^\text{train}$ and testing set $\mathcal{T}^\text{test}$, all tasks with an angles greater than $1.5 \cdot \pi$ (independently of the radius) are treated as testing task and are not part of the training set.
This split yields 48 training tasks, and 12 testing tasks. 
Spatially, the test set is indicated by the shaded are in Figure~\ref{fig:spatio-temporal}.

\textbf{MuJoCo Half Cheetah:} This task involves continuous joint control of a planar half-cheetah agent. The state and action spaces have dimensions 20 and 7, respectively. Tasks vary by target velocity, and the reward is proportional to the deviation from this target. Although the target velocity is not part of the observable state, it is implicitly encoded via the reward, ensuring that each task defines a fully observable MDP. There are 40 tasks in total, 35 are used for training and 5 are used for testing. For more environment details see~\citep{duan2016benchmarking, xu2022prompting}.

\textbf{MuJoCo Ant:} As in~\citet{xu2022prompting}, this task requires continuous control of an ant agent via eight joint actuators. The 27-dimensional state space includes positions and velocities of the agent’s body. Tasks involve moving in different directions, with reward based on velocity along the target heading. There are 50 tasks in total, 45 are used for training and 5 are used for testing. For more environment details see~\cite{duan2016benchmarking, xu2022prompting}.

\textbf{Meta-World Pick-Place:} Following~\citet{yu2020meta}, this task involves Cartesian control of a simulated Sawyer robot’s end-effector. The 4-dimensional action space includes 3D position deltas and gripper torque. The 39-dimensional state space encodes gripper and object positions, and quaternions, but excludes the goal location. The benchmark consists of 50 task variations with different placement goals; 5 are held out during training for evaluation.
\clearpage

\section{Training Details}
\label{appx:training_details}

All experiments were conducted on an instance equipped with NVIDIA T4 GPU with 8GB of memory, utilizing the PyTorch library~\citep{paszke2019pytorch}. Pre-training the multi-task PDT backbone took approximately 5 hours. Performing 250 online rollouts and prompt tuning took approximately 30 to 120 minutes per task, depending on environment and algorithm.

Details of the hyperparameters for the DT and PDT models are provided in Table~\ref{tab:dt_hyperparameters}, while those for the bandit model are listed in Table~\ref{tab:cmab_hyperparameters}. 
We used standard values for each hyperparameter instead of performing an expensive hyperparameter optimization.
Environment-specific hyperparameters are reported separately in Table~\ref{tab:pdt_hyperparameters_env}. The implementations of DT and PDT were adopted from \citet{minimal_decision_transformer} with minimal modifications to integrate seamlessly with the CMAB prompt-tuning framework.
The MuJoCo experiments were performed by integrating our prompt-tuning bandits into the official PDT repository~\cite{xu2022prompting}.
For the hill-climbing and ZORankSGD~\citep{hu2023prompt} baselines, we sample an initial prompt randomly from the expert demonstration dataset for the target task, then apply Gaussian noise and perform hill climbing. To encourage convergence to a final prompt, we linearly anneal the scale of the exploration noise from 1.1 to 0.1 over the course of the 250 online rollouts.

\begin{table*}[h]
\begin{minipage}{0.48\linewidth}
    \begin{center}
        \begin{tabular}{ll}
            \multicolumn{1}{c}{Hyperparameter}  & \multicolumn{1}{c}{Value} \\
            \hline \\
            Number of transformer blocks & 3\\
            Number of attention heads & 1 \\
            Embedding dimension & 128 \\
            Transformer activation function & GELU \\
            MLP activation function & ReLU \\
            MLP hidden layers & 2 \\
            MLP width & 128 \\
            Batch size (per task) & 8\\
            Learning rate & 1e-4\\
            Learning rate decay weight & 1e-4\\
            Optimizer & AdamW \\
        \end{tabular}
    \end{center}
    \caption{Hyperparameter values for PDT and DT models.}
    \label{tab:dt_hyperparameters}
\end{minipage}%
\hfill
\begin{minipage}{0.48\linewidth}
    \begin{center}
        \begin{tabular}{ll}
            \multicolumn{1}{c}{Hyperparameter}  & \multicolumn{1}{c}{Value} \\
            \hline \\
            Batch size & All data\\
            Learning rate & 1e-3\\
            Evaluation trials & 250\\
            $\epsilon$ in $\epsilon$-greedy & 0.1\\
            $c$ (exploration parameter) in UCB & 3\\
            Reward model type & MLP \\
            MLP hidden layers & 2 \\
            MLP width & 16 \\
            MLP activation function & ReLU \\
            Optimizer & Adam \\
        \end{tabular}
    \end{center}
    \caption{Hyperparameters of Contextual Multi-armed Bandit}
    \label{tab:cmab_hyperparameters}
\end{minipage}
\end{table*}

\begin{table*}[ht]
    \begin{center}
    \begin{tabular}{llll}
    \multicolumn{1}{c}{Environment}  &\multicolumn{1}{c}{Target Return} &\multicolumn{1}{c}{Prompt Length $(J \times H)$ } &\multicolumn{1}{c}{Context Length}
    \\ \hline \\
    2D-Point-Sparse & 10 & $(1 \times 3), (2 \times 3), (4 \times 3)$ & 5\\
    Cheetah-vel & 0 & $(1 \times 5), (2 \times 20)$ & 20 \\
    \end{tabular}
    \end{center}
    \caption{Environment-specific hyperparameters of DT and PDT.}
    \label{tab:pdt_hyperparameters_env}
\end{table*}

\section{Additional Results}
\label{appx:additional_plots}
\subsection{Online Rollouts in \texttt{Sparse 2D point}}\label{app:plots_2d_rollouts}
\begin{figure}[H]
    \centering
    \begin{subfigure}{\linewidth}
        \centering
        \includegraphics[width=\linewidth]{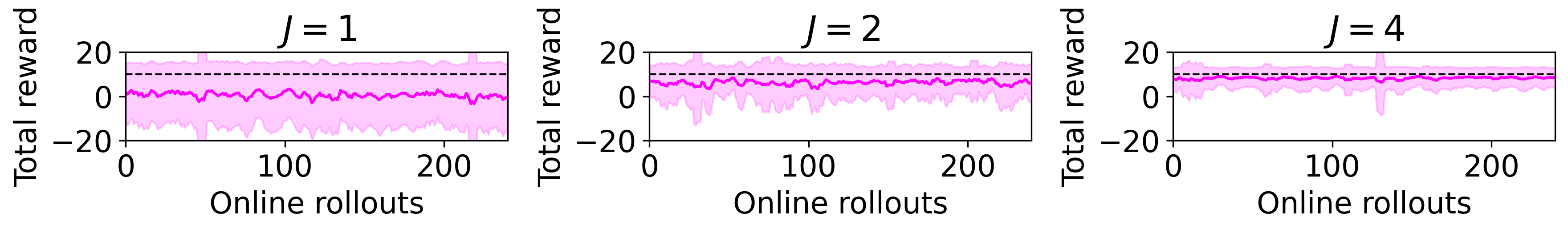}
        \caption{Online performance of \textbf{standard PDT, without prompt-tuning}.}
        \label{fig:standardPDT_onlineRollouts_2d}
    \end{subfigure}
    \vspace{0.5em}
    \begin{subfigure}{\linewidth}
        \centering
        \includegraphics[width=\linewidth]{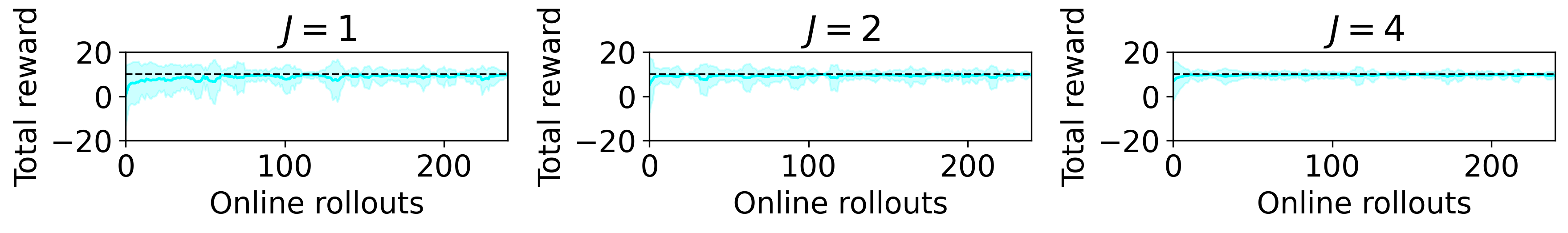}
        \caption{Online performance \textbf{with prompt-tuning}, using \textbf{$\epsilon$-greedy exploration} and \textbf{transformer features $\phi$} as segment representation for the bandit's reward models.}
        \label{fig:epsgreedy_features_onlineRollouts_2d}
    \end{subfigure}
    \begin{subfigure}{\linewidth}
        \centering
        \includegraphics[width=\linewidth]{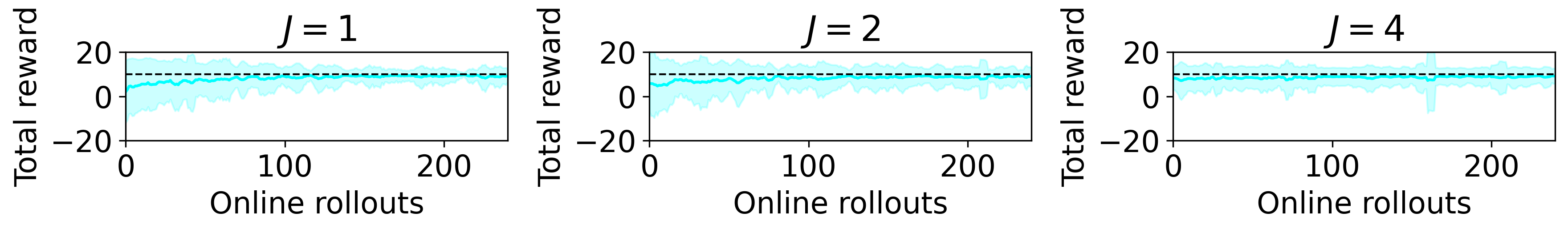}
        \caption{Online performance \textbf{with prompt-tuning}, using \textbf{UCB exploration} and \textbf{transformer features $\phi$} as segment representation for the bandit's reward models.}
        \label{fig:ucb_features_onlineRollouts_2d}
    \end{subfigure}
    \begin{subfigure}{\linewidth}
        \centering
        \includegraphics[width=\linewidth]{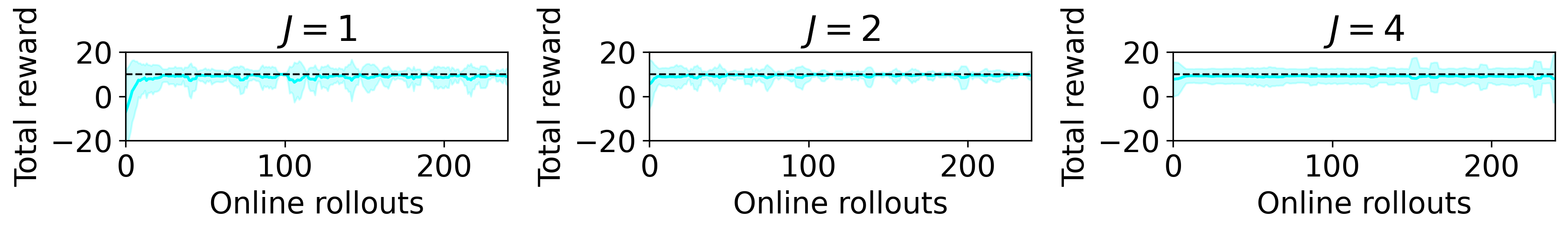}
        \caption{Online performance \textbf{with prompt-tuning}, using \textbf{$\epsilon$-greedy exploration} and \textbf{unencoded trajectory segments} instead of transformer features for the bandit's reward models.}
        \label{fig:epsgreedy_rawSegments_onlineRollouts_2d}
    \end{subfigure}
    \begin{subfigure}{\linewidth}
        \centering
        \includegraphics[width=\linewidth]{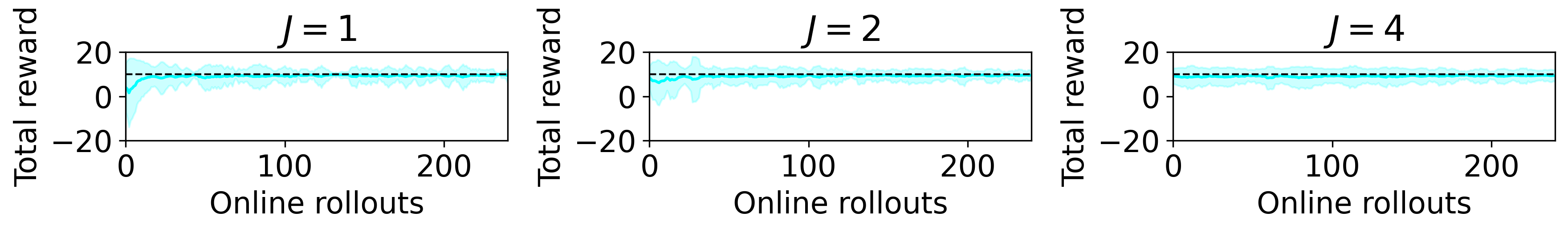}
        \caption{Online performance \textbf{with prompt-tuning}, using \textbf{UCB exploration} and \textbf{unencoded trajectory segments} instead of transformer features for the bandit's reward models.}
        \label{fig:ucb_rawSegments_onlineRollouts_2d}
    \end{subfigure}
    \begin{subfigure}{\linewidth}
        \centering
        \includegraphics[width=\linewidth]{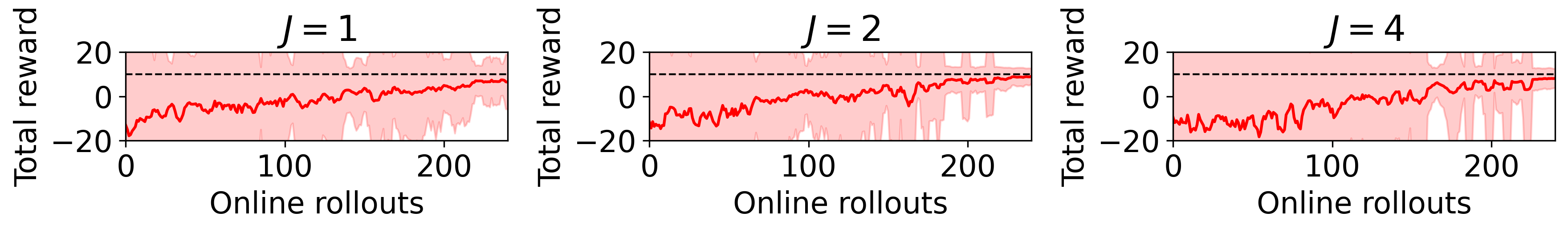}
        \caption{Online performance \textbf{with prompt-tuning}, using \textbf{Gaussian noise} and \textbf{hill climbing} to optimize prompt selection.}
        \label{fig:hillclimbing_onlineRollouts_2d}
    \end{subfigure}
    \caption{Plots of data used in Table~\ref{tab:2D-id-results}. The dashed line marks the optimal return at +10, the shaded area corresponds to 1 standard deviation around the mean. }
    \label{fig:all_prompt_tuning_results}
\end{figure}
\begin{figure}[H]
    \centering
    \begin{subfigure}{\linewidth}
        \centering
        \includegraphics[width=\linewidth]{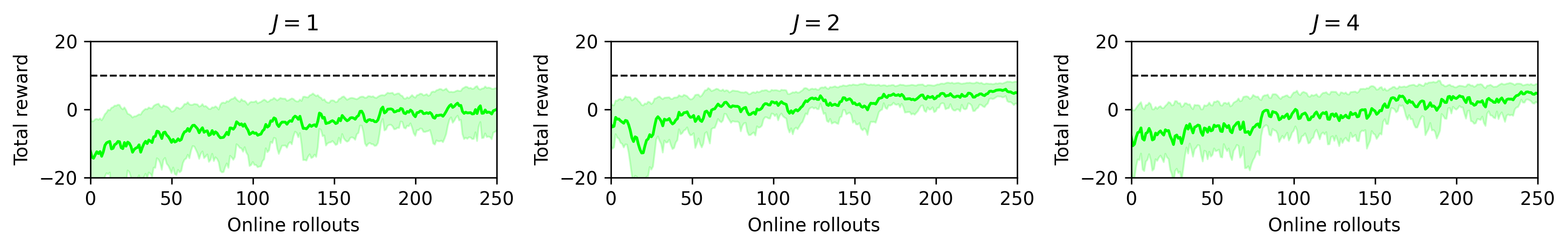}
        \caption{Online performance after 250 episode \textbf{with prompt-tuning}, using \textbf{ZORankSGD} for prompt tuning.}
        \label{fig:zoranksgd_250_onlineRollouts_2d}
    \end{subfigure}
    \begin{subfigure}{\linewidth}
        \centering
        \includegraphics[width=\linewidth]{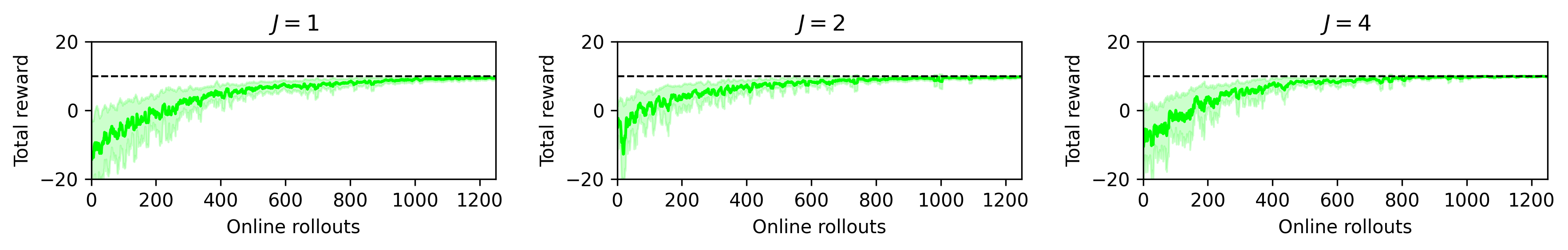}
        \caption{Online performance after 1500 episodes \textbf{with prompt-tuning}, using \textbf{ZORankSGD} for prompt tuning.}
        \label{fig:zoranksgd_1500_onlineRollouts_2d}
    \end{subfigure}
    \caption{Plots of ZORankSGD prompt-tuning as reported in Table~\ref{tab:2D-id-results}. The dashed line marks the optimal return at +10. The shaded area corresponds to one standard deviation around the mean. ZORankSGD requires the online evaluation of $m = 5$ prompt perturbations to compute one prompt update. Results using a budget of 250 online rollouts (and 50 prompt improvement steps) are shown in subfig (a). Results using 1250 online rollouts (and 250 prompt improvements) are shown in subfig (b).}
    \label{fig:all_prompt_tuning_results}
\end{figure}

\clearpage

\subsection{Prompt Quality Experiment for the \texttt{Sparse 2D point} Environment}\label{app:plots_prompt_quality}

\textbf{Can prompt-tuning exploit non-expert datasets?}
While the PDT model with a random prompt selection strategy relies on access to expert demonstrations, our method demonstrates greater robustness to the quality of demonstration data. To examine this, we generate additional prompt datasets $\{\mathcal{P}_i^{\%j}\}, j \in \{0, 10, \dots, 100\}$, which combine $j\%$ of expert data along with novice demonstrations, selected from trajectories in the bottom 5th percentile of task returns (see Figure~\ref{subfig:expert_vs_novice}). 

When sampling prompts from these mixed datasets, the bandit-based prompt optimization significantly enhances the PDT model's robustness and improves its performance, as shown in Table~\ref{tab:2d-prompt-quality-exp}. This approach reduces reliance on pure expert demonstrations by learning to identify the optimal prompt from any arbitrary mixture dataset.
\begin{table}[H]
\centering
\begin{tabular}{cccc}
\makecell{Expert percentage $j\%$} & No tuning & $\epsilon$-greedy & UCB\\
\hline
0\% & -39.4 $\pm$ 16.8 & -12.1 $\pm$ 19.4 & -3.8 $\pm$ 10.0 \\
10\% & -40.8 $\pm$ 14.1 & 4.4 $\pm$ 3.4 & 7.3 $\pm$ 1.9 \\
20\% & -49.6 $\pm$ 32.8 & 5.5 $\pm$ 3.4 & 9.4 $\pm$ 0.5 \\
30\% & -50.1 $\pm$ 27.4 & 6.5 $\pm$ 1.3 & \textbf{9.8 $\pm$ 0.3} \\
40\% & -41.4 $\pm$ 26.4 & 2.1 $\pm$ 6.4 & 8.6 $\pm$ 2.0 \\
50\% & -12.9 $\pm$ 4.0 & 5.1 $\pm$ 4.4 & 8.7 $\pm$ 0.8 \\
60\% & -14.1 $\pm$ 5.0 & 7.6 $\pm$ 1.5 & 8.5 $\pm$ 1.2 \\
70\% & -28.6 $\pm$ 17.0 & 8.4 $\pm$ 0.9 & 7.7 $\pm$ 1.3 \\
80\% & -12.3 $\pm$ 11.1 & 8.6 $\pm$ 1.0 & \textbf{9.8 $\pm$ 0.3} \\
90\% & {0.9 $\pm$ 0.5} & 8.6 $\pm$ 0.9 & 8.6 $\pm$ 1.0 \\
100\% & \textbf{0.7 $\pm$ 1.7} & \textbf{9.7 $\pm$ 0.4} & 9.7 $\pm$ 0.4 \\
\hline
\end{tabular}
\caption{
Without prompt-tuning, PDT's performance deteriorates with the percentage of expert trajectories in $\mathcal{P}_i$. 
Our prompt-tuning method is robust with respect to the percentage of expert data and achieves near-optimal performance with as little as 10\% expert demonstrations. Results are averaged over three seeds for a single training task.}
\label{tab:2d-prompt-quality-exp}
\end{table}

\begin{figure*}[h!]
    \centering
    \includegraphics[width=0.55\linewidth]{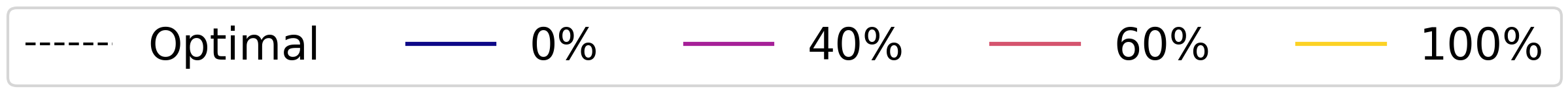}
    \vfill
    \begin{subfigure}{0.3\textwidth}
        \includegraphics[width=\linewidth]{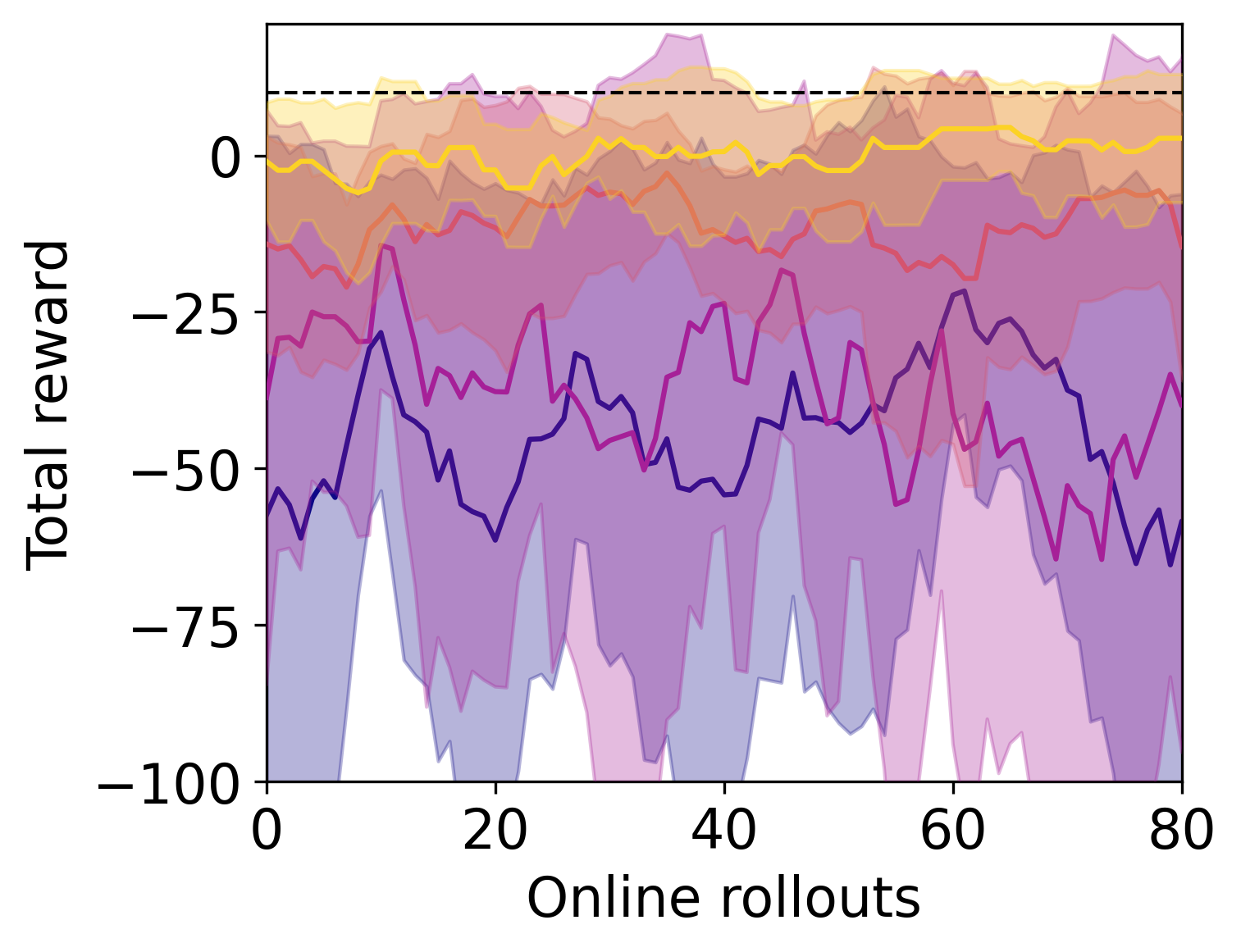}
        \caption{Baseline PDT, no tuning}
        \label{fig:mixture-no-tuning}
    \end{subfigure}
    \hfill
    \begin{subfigure}{0.3\textwidth}
        \includegraphics[width=\linewidth]{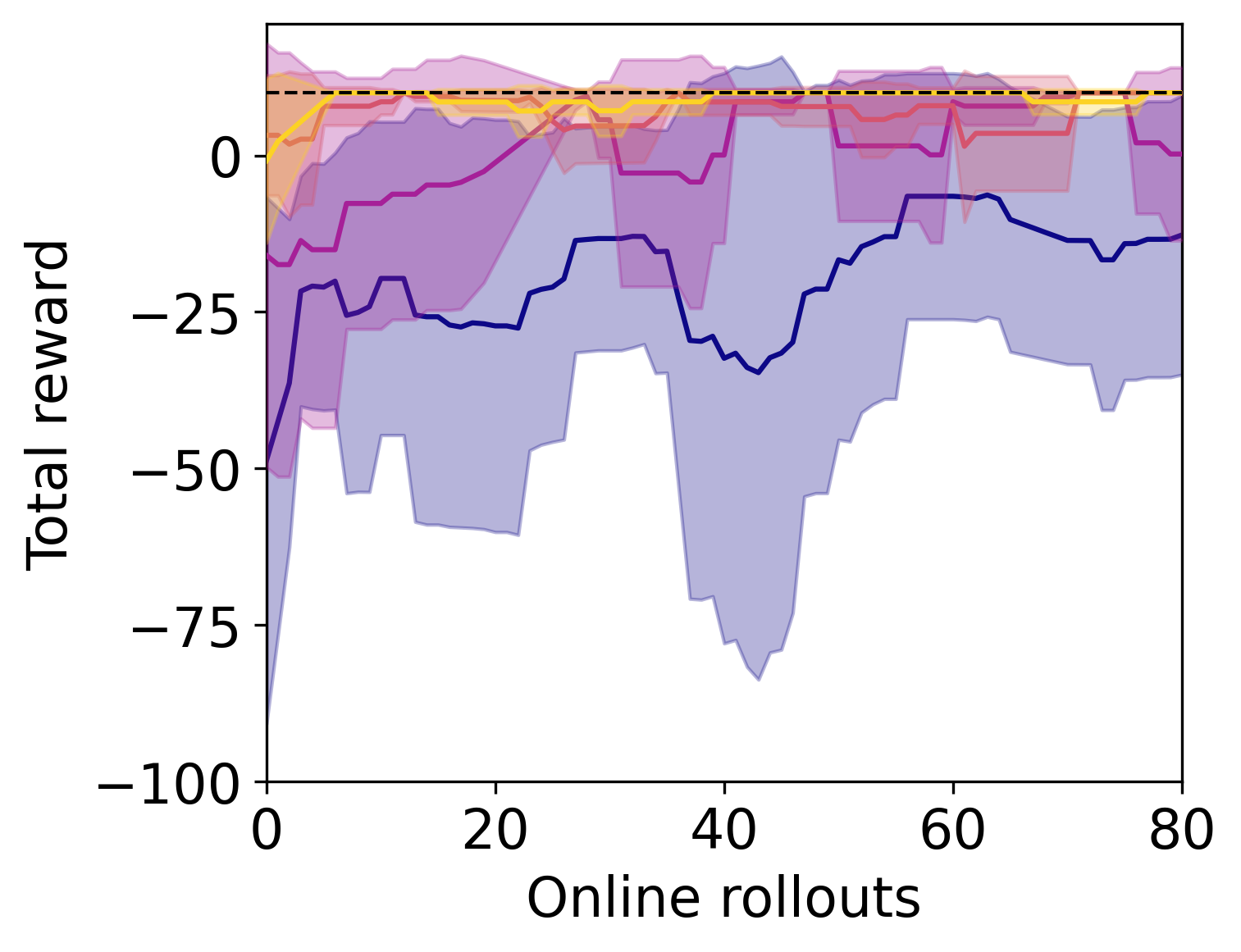}
        \caption{$\epsilon$-greedy prompt tuning}
        \label{fig:mixture-eps-greedy}
    \end{subfigure}
        \hfill
    \begin{subfigure}{0.3\textwidth}
        \includegraphics[width=\linewidth]{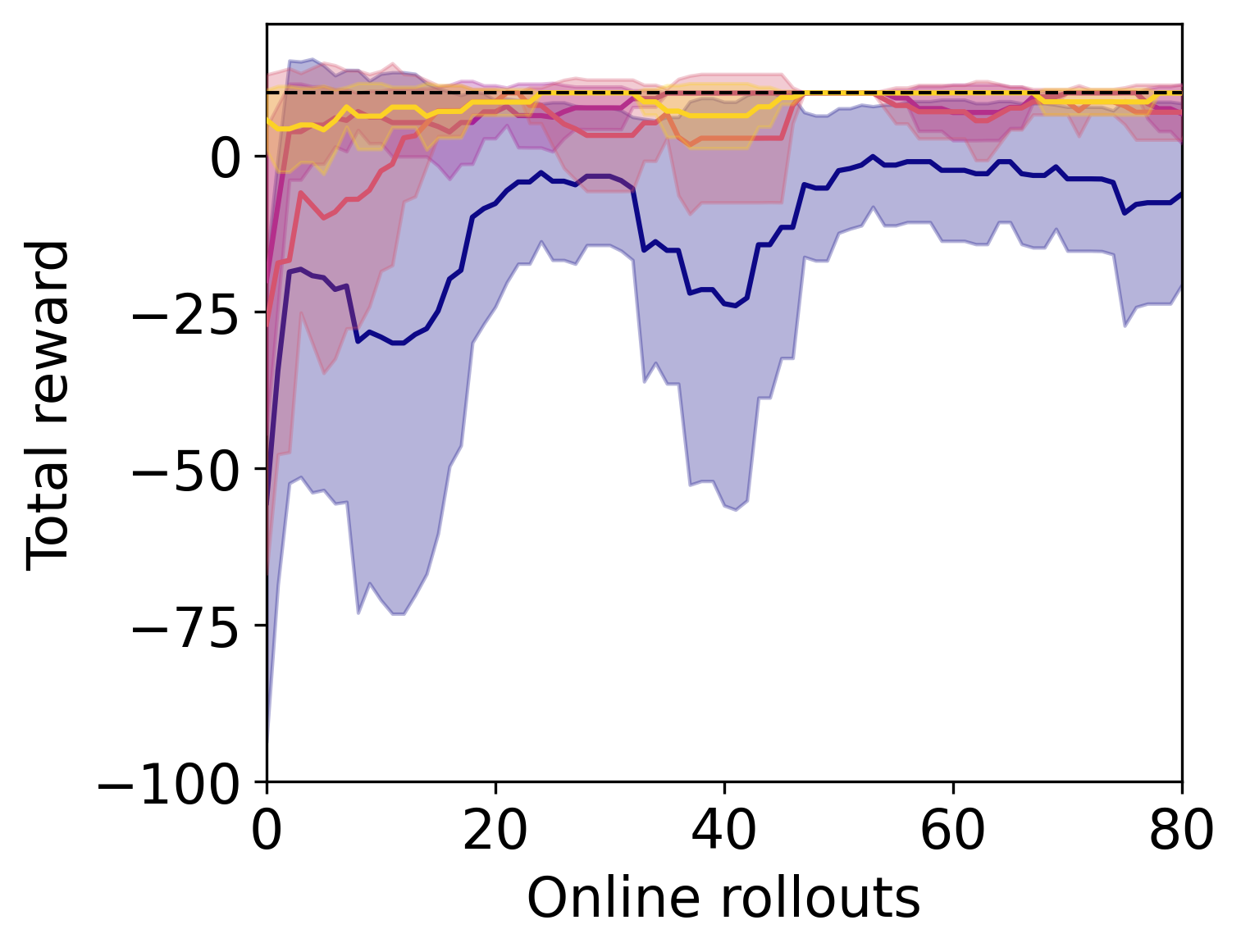}
        \caption{UCB prompt tuning}
        \label{fig:mixture-ucb}
    \end{subfigure}
    \caption{Prompt quality experiment. Color indicates percentage of expert demonstrations in $\mathcal{P}_i$. Without prompt-tuning, PDT's performance degrades and is roughly proportional to the percentage of expert demonstrations in the dataset. Our prompt-tuning method quickly recovers and converges to near-optimal performance by finding the high-performance prompts in the mixture dataset. The shaded region denotes 1 standard deviation around the mean, averaged over three random seeds for a single training task.}
    \label{fig:results-mixture-exp}
\end{figure*}
\clearpage

\clearpage
\section{Segment Independence Attention Analysis}\label{app:segment_attention}
The key assumption based on which we design our structured bandit architecture is that prompt segments contribute independently and additively to prompt informativeness. 
This assumption implies that PDT can identify the target tasks by attending to key $(\hat{r}, \mathbf{s}, \mathbf{a})$-pairs in the prompt, as opposed to deriving task identity from the interaction effects between prompt segments, or by attending to the prompt globally. 
This assumption is crucial, because it allows us to decompose the prompt tuning problem into a bandit that maintains $J$ reward models, one for each prompt segment, which results in a reduction of the size of the exploration problem from combinatorial in $J$ to linear in $J$. 
The basis for the assumption is that a) many MDPs can be characterized by their optimal state-action marginal (i.e. by key $(\mathbf{s}, \mathbf{a})$-pairs) and b) that PDT's pre-training paradigm discourages reliance on segment interaction or arrangement, since prompt segments are sampled uniformly at random during pre-training, not by enforcing order or other constraints between segments. 
This training setup implicitly encourages invariance to permutation and inter-segment dependencies, and thus we believe that this limits the influence of global prompt structure on the model's behavior.

\begin{figure}
     \centering
     \begin{subfigure}[b]{0.78\textwidth}
         \centering
         \includegraphics[width=\textwidth]{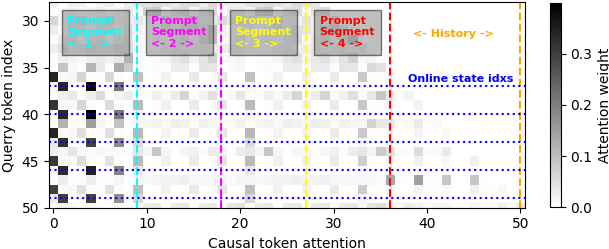}
         \caption{Prompt segment 1 is informative.}
         \label{subfig:seg_1_informative}
     \end{subfigure}
     \hfill
     \begin{subfigure}[b]{0.78\textwidth}
         \centering
         \includegraphics[width=\textwidth]{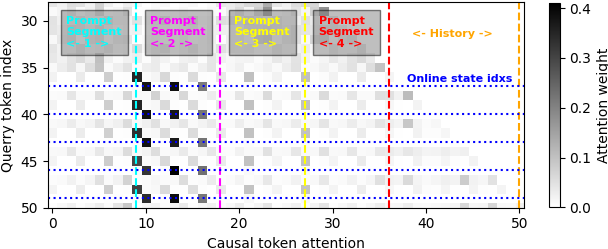}
         \caption{Prompt segment 2 is informative.}
         \label{subfig:seg_2_informative}
     \end{subfigure}
     \begin{subfigure}[b]{0.78\textwidth}
         \centering
         \includegraphics[width=\textwidth]{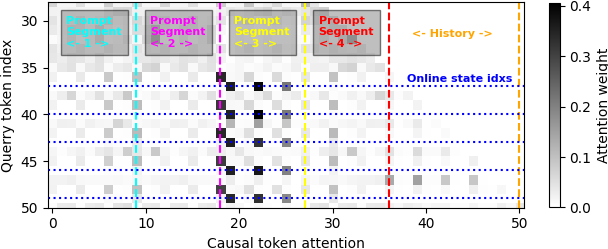}
         \caption{Prompt segment 3 is informative.}
         \label{subfig:seg_3_informative}
     \end{subfigure}
     \hfill
     \begin{subfigure}[b]{0.78\textwidth}
         \centering
         \includegraphics[width=\textwidth]{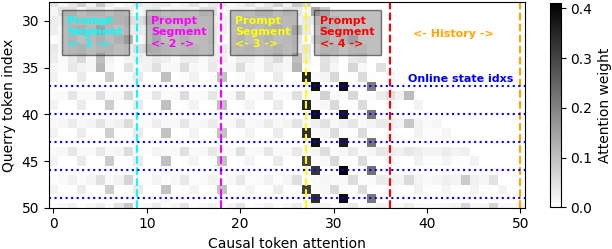}
         \caption{Prompt segment 4 is informative.}
         \label{subfig:sed_4_informative}
     \end{subfigure}
        \caption{Attention weights from the first transformer block (after softmax), darker values indicate higher attention scores. The first 20 rows have been cropped to improve readability. The blue horizontal lines indicate the indices of state tokens at times $t-H, t-H+1, \dots, t$ from which the action at the corresponds $t$ is predicted. It can be seen that PDT attends mostly to the informative prompt token, independently of it's position or neighboring segments.}
        \label{fig:PDT-segment_attention}
\end{figure}

To provide additional support for this interpretation, we conduct an analysis of PDT's attention weight on our 2D environment. 
Here, for each task, we construct prompts that contain exactly one segment that is highly informative for task identification, based on our domain knowledge, while the remaining $J - 1$ segments carry little information for discriminating between tasks. 
We then performed rollouts using different permutations of these segments as prompts. 
The performance of PDT remained robust under those permutations with $6.27 \pm 0.34$ return on average over tasks, prompts, and model configurations ($J=2, H=3$ and $J=4, H=3$), which implies prompt segment permutation invariance in PDT. 
We visualize representative PDT attention mask for different prompt permutations in Figure~\ref{fig:PDT-segment_attention} to confirm qualitative that PDT indeed identifies tasks by attending mostly to the individual, informative prompt segment.

Furthermore, we quantify attention to each prompt segment by computing the mean of the token attention score of tokens in each prompt segment. We then sum the per-timestep mean segment attention scores, and find that in 99\% of rollouts conducted during this analysis, PDT assigned highest attention to the informative segment. Additionally, we manually set attention weights for specific prompt segments to zero, and find that PDT's performance remains stable when uninformative segments are masked out, while it degrades drastically when the informative segment is removed, as per Table~\ref{tab:2d-attention-masking-result}.

\begin{table}[!ht]
    \centering
    \begin{tabular}{|c|c|c|}
        \hline
        Full prompt & Only informative segment & Only uninformative segments \\ 
        \hline
        $6.27 \pm 0.34$ & $5.88 \pm 0.23$ & $-34.06 \pm 4.49$ \\
        \hline
    \end{tabular}
    \caption{PDT performance with different attention scopes.}
\label{tab:2d-attention-masking-result}
\end{table}

These results suggest that PDT behavior is largely driven by the most informative segment, and that inter-segment interactions are negligible in practice, which supports and justifies our assumption and bandit architecture.
However, we also observed instances where PDT attended to all prompt segments rather uniformly, despite $J-1$ segments not being useful for task discrimination as per our domain knowledge. We hypothesize that these instances were due to slight overfitting to the prompt dataset and see the design of regularization that prevents PDT from exploit such nuances in the prompt dataset as important future work.

\clearpage
\section{Wall-clock inference time}\label{app:wall-clock-time}
Table~\ref{tab:wall-clock-times} reports representative wall-clock inference times for different methods on the \texttt{MuJoCo Ant} environment. 
Comparing our bandit architecture with PDT-encoded prompt representations (denoted by $^\Psi$) to its unencoded counterpart highlights the efficiency benefits of using the PDT as a feature extractor. 
When operating on PDT-encoded prompts, the Thompson Sampling bandit maintains low inference times even for large prompts $(J=2, H=20)$, since the PDT-encoded representation has a fixed dimension $\mathbb{R}^d$ determined by the PDT's token embedding space, independent of the prompt length. 
In contrast, unencoded prompts lead to input sizes that scale as $H \times (|\mathcal{S}| + |\mathcal{A}| + 1)$ for each segment, substantially increasing inference time due to the higher cost of updating the reward models.
\begin{table}[]
    \centering
    \begin{tabular}{lcc}
        Method & \texttt{MuJoCo Ant} ($J=1, H=5$) & \texttt{MuJoCo Ant} ($J=2$, H=20) \\ 
        \hline
        PDT (full-model fine-tuning) & $\sim28$m & $\sim 38$m \\
        Hill climbing & $\sim16$m & $\sim 24$m \\
        ZORankSGD & $\sim 52$m &  $\ \ \sim 120$m \\
        TS$^\Psi$ & $\sim 28$m & $\sim 34$m \\
        TS & $\sim 33$m & $\ \ \sim 172$m \\
        \hline
    \end{tabular}
    \caption{Wall-clock inference times with different methods.}
    \label{tab:wall-clock-times}
\end{table}

\section{Proof of Theorem~\ref{thm:regret_bound}}
\label{appx:regret_proof}
\begin{theorem*}
Assume that the reward function $G\colon P^J \to \mathbb{R}$ for a prompt $\rho = (\tilde{\tau}_1,\dots,\tilde{\tau}_J)$ decomposes as the mean of $J$ independent reward models $\phi_j(\tilde{\tau}_j)$:
\begin{equation*}
    G(\rho) = \frac{1}{J} \sum_{j=1}^{J} \phi_j(\tilde{\tau}_j) + h(\tilde{\tau}_1,\dots,\tilde{\tau}_J),
\end{equation*}
and that the interaction term is uniformly bounded by
$
    |h(\tilde{\tau}_1,\dots,\tilde{\tau}_J)| \leq \varepsilon,\quad \forall\, \tilde{\tau}_j \in P.
$
Let $\rho^* = (\tilde{\tau}^*_1,\dots,\tilde{\tau}^*_J)$ denote the optimal prompt, and suppose that for each slot $j$, a bandit algorithm guarantees a slot-specific regret
\begin{align*}
\mathrm{Regret}_j(K) = \sum_{t=1}^{K} \mathbb{E}\left[\phi_j(\tilde{\tau}^*_j)-\phi_j(\tilde{\tau}_{t,j})\right]
\end{align*}
over $K$ rounds. Then the cumulative regret after $K$ rounds is bounded as:
\begin{align*}
    \mathrm{Regret}(K) &\triangleq \sum_{t=1}^{K} \mathbb{E}\left[G(\rho^*) - G(\rho_t)\right] %
    \leq \frac{1}{J} \sum_{j=1}^{J} \mathrm{Regret}_j(K) + 2K\varepsilon. \label{eq:regret_bound}
\end{align*}
\end{theorem*}
\begin{proof}
We begin with the decomposition of the reward function. The instantaneous regret at round $t$ is
\[
G(\rho^*) - G(\rho_t) = \left(\frac{1}{J} \sum_{j=1}^{J} \phi_j(\tau_j^*) + h(\tau_1^*,\dots,\tau_J^*)\right) - \left(\frac{1}{J} \sum_{j=1}^{J} \phi_j(\tau_{t,j}) + h(\tau_{t,1},\dots,\tau_{t,J})\right).
\]
Rearranging terms yields:
\[
G(\rho^*) - G(\rho_t) = \frac{1}{J} \sum_{j=1}^{J} \Bigl(\phi_j(\tau_j^*) - \phi_j(\tau_{t,j})\Bigr) + \Bigl[h(\tau_1^*,\dots,\tau_J^*) - h(\tau_{t,1},\dots,\tau_{t,J})\Bigr].
\]
By the bounded interaction assumption, we have:
\[
\Bigl|h(\tau_1^*,\dots,\tau_J^*) - h(\tau_{t,1},\dots,\tau_{t,J})\Bigr| \leq 2\varepsilon.
\]
Thus, taking expectations and summing over $t=1$ to $K$, the cumulative expected regret satisfies
\begin{align*}
\mathrm{Regret}(K) &= \sum_{t=1}^{K} \mathbb{E}\left[G(\rho^*) - G(\rho_t)\right] \\
&\leq \sum_{t=1}^{K} \left[\frac{1}{J} \sum_{j=1}^{J}\, \mathbb{E}\Bigl(\phi_j(\tau_j^*) - \phi_j(\tau_{t,j})\Bigr) + 2\varepsilon\right] \\
&=\frac{1}{J} \sum_{j=1}^{J} \sum_{t=1}^{K} \mathbb{E}\left[\phi_j(\tau_j^*) - \phi_j(\tau_{t,j})\right] + 2K\varepsilon.
\end{align*}
By definition, the slot-specific cumulative regret is given by
\[
\mathrm{Regret}_j(K) = \sum_{t=1}^{K} \mathbb{E}\left[\phi_j(\tau_j^*) - \phi_j(\tau_{t,j})\right],
\]
so we obtain the final bound:
\[
\mathrm{Regret}(K) \leq \frac{1}{J} \sum_{j=1}^{J} \, \mathrm{Regret}_j(K) + 2K\varepsilon.
\]
\end{proof}

\clearpage
\newpage

\end{document}